# Navigator-free EPI Ghost Correction with Structured Low-Rank Matrix Models: New Theory and Methods

Rodrigo A. Lobos, *Student Member, IEEE*, Tae Hyung Kim, *Student Member, IEEE*,
W. Scott Hoge, *Member, IEEE,* and Justin P. Haldar, *Senior Member, IEEE*

*Abstract*—Structured low-rank matrix models have previously been introduced to enable calibrationless MR image reconstruction from sub-Nyquist data, and such ideas have recently been extended to enable navigator-free echo-planar imaging (EPI) ghost correction. This paper presents novel theoretical analysis which shows that, because of uniform subsampling, the structured low-rank matrix optimization problems for EPI data will always have either undesirable or non-unique solutions in the absence of additional constraints. This theory leads us to recommend and investigate problem formulations for navigator-free EPI that incorporate side information from either image-domain or k-space domain parallel imaging methods. The importance of using nonconvex low-rank matrix regularization is also identified. We demonstrate using phantom and *in vivo* data that the proposed methods are able to eliminate ghost artifacts for several navigator-free EPI acquisition schemes, obtaining better performance in comparison to state-of-the-art methods across a range of different scenarios. Results are shown for both single-channel acquisition and highly accelerated multi-channel acquisition.

## I. INTRODUCTION

ECHO-planar imaging (EPI) [1] is currently one of the fastest MRI pulse sequences and one of the most popular sequences for functional, diffusion, and perfusion imaging. EPI uses a train of gradient echoes to measure multiple lines of k-space from a single excitation, but is prone to artifacts because it employs a long readout, uses rapidly-switching high-amplitude gradients, and measures alternating lines of k-space with different gradient polarities [2].

In conventional single-shot EPI, even and odd lines of k-space are acquired with alternating gradient polarities. In practice, hardware imperfections, eddy currents, field inhomogeneity, concomitant fields, system delays, and similar phenomena can introduce signal phase errors between k-space lines acquired with different readout gradient polarities. If these phase errors are not correctly compensated, a Nyquist (or $N/2$) ghost artifact is observed corresponding to an aliased image that is positioned a half field-of-view (FOV) away from the true spatial position along the phase-encoding direction. In multi-shot EPI, full k-space coverage is achieved by using multiple excitations, where a different segment of k-space is acquired using EPI for each shot. Multi-shot EPI is used to reduce the EPI echo train, which subsequently reduces distortion and spin-dephasing effects from local field inhomogeneity. Images can then be reconstructed by interleaving the multi-shot data together. As in the single-shot case, the mismatch between different gradient polarities also leads to Nyquist ghost artifacts for multi-shot data. However, multi-shot data may also exhibit additional ghost artifacts if there happen to be inconsistencies between each shot – including system drift, subject motion, and, particularly in the case of gradient-recalled EPI, respiration.

Many approaches to ghost correction have been proposed over the years, which we group into two main categories. The first category contains simple model-based approaches such as Refs. [3]–[10], which assume a low-dimensional model to describe a systematic phase mismatch between even and odd lines. The parameters of the mismatch model are often estimated using separate navigator data, and can then be used to correct the mismatch in the measured EPI data. While these methods are widely used and can work well when the mismatch model is accurate, phenomena such as eddy currents can lead to more complicated data mismatches that are not fully captured by simple models.

This paper focuses on the second category of methods, which includes those described in Refs. [11]–[15]. These methods rely on a more flexible model in which the data samples of each gradient polarity/shot are assumed to be coming from different but highly-correlated images. For example, it is often assumed that the images corresponding to different gradient polarities or shots have the same image magnitudes but different image phases.[1] This is similar to how parallel imaging methods like SENSE [16] and GRAPPA [17] assume that the different channels of an array receiver coil acquire images that are different (i.e., modulated by different coil sensitivity profiles) but highly correlated. As a result, it is not

R. Lobos, T.H. Kim and J.Haldar are with the Signal and Image Processing Institute, Ming Hsieh Department of Electrical Engineering, University of Southern California, Los Angeles, CA, 90089, USA.
W.S. Hoge is with the Brighman and Women's Hospital, Radiology Department, and Harvard Medical School, Radiology Department, Boston, MA, USA.
This work was supported in part by NSF grant CCF-1350563 and NIH grants R21-EB022951, R01-NS074980, and R01-NS089212.

---

[1]This assumption is a simplification of the true imaging physics, and may not fully account for any time-dependent effects that evolve dynamically during data acquisition (e.g., due to eddy currents). More detailed modeling has been considered in some previous work, e.g., [3], though the simplified image-domain model described above underlies much of the recent literature.

surprising that many recent ghost correction approaches can be viewed as adaptations of previous parallel imaging methods to the ghost correction context. An example is the dual-polarity GRAPPA (DPG) method [13], which treats different polarities as if they were different virtual coils, and uses a dual GRAPPA kernel (with the GRAPPA weights divided into two halves corresponding to the two different gradient polarities) to synthesize a ghost-free fully-sampled image. Even though methods from the second category have been shown to have state-of-the-art performance in many challenging scenarios, they can still suffer from artifacts in certain cases. For example, DPG can fail to successfully correct ghost artifacts if there are mismatches between the measured EPI data and the autocalibration signal (ACS) used to train the dual GRAPPA kernel. This type of mismatch can occur because of changes in the measured data as a function of time, e.g., due to respiration [18]. In addition, most of the methods in the second category rely on the use of multichannel data, and are not easily applicable to single-channel ghost-correction.

Recently, novel image reconstruction methods have been proposed that enable calibrationless single-channel and multi-channel image reconstruction from undersampled k-space data using structured low-rank matrix (SLM) completion approaches [19]–[26]. SLM approaches are based on the assumption that there exist linear dependencies in k-space due to limited image support, smooth image phase variations, parallel imaging constraints, and/or transform-domain image sparsity. While such constraints have been used before in EPI ghost correction, e.g., Refs. [3], [27], the SLM formulation of these constraints is distinct from classical approaches. SLM approaches were not originally developed for EPI data, but have very recently been adapted to such contexts [24], [28]–[31]. These approaches have demonstrated to yield state-of-the-art performance in highly-accelerated EPI image reconstruction [24] and the ability to perform navigator-free EPI ghost correction [28]–[31].

In this paper, we analyze theoretical aspects of navigator-free EPI ghost correction using SLM approaches and obtain new insights that have major implications for ghost correction performance. Specifically, we prove that the SLM completion problem associated with ghost correction either has a non-unique solution or a unique solution that is undesirable. Based on this result, we observe that constraints are needed to ensure the performance of SLM-based ghost correction, and investigate two approaches that achieve substantially improved results. A preliminary account of portions of this work was previously given in Ref. [32].

In our first approach, we combine ideas from the LORAKS [20]–[22], [24] SLM framework with coil sensitivity maps within the SENSE framework, as has previously been done for EPI reconstruction [24] and EPI ghost correction [29], [30]. Compared to MUSSELS [29], [30] (a similar SENSE-based ghost correction method), our new approach makes use of a nonconvex regularization function from earlier LORAKS work [20]–[24] which yields improved results, both in theory and in practice, than the convex approach used in MUSSELS. Additionally, MUSSELS uses one of the simpler forms of SLM construction (named the **C**-matrix in the terminology of LORAKS [20], [22]) that incorporates support and parallel imaging constraints, but does not leverage image phase constraints. Our new SENSE-based approach takes advantage of a more advanced SLM construction (named the **S**-matrix in the terminology of LORAKS [20], [22]).

Our second approach combines LORAKS with k-space domain parallel imaging linear predictability constraints, like those used in GRAPPA [17], SPIRiT [33], and PRUNO [34]. In our implementation, these constraints are imposed within the broader framework of autocalibrated LORAKS (AC-LORAKS) [35]. To the best of our knowledge, this is the first time that this type of information has been combined with structured low-rank matrix completion methods in the context of EPI ghost correction. This second new approach not only works for multi-channel data as expected, but remarkably, we observe it also works for ghost correction of single-channel data in some cases.

This paper is organized as follows. Section II reviews SLM approaches and defines the notation used in the rest of the paper. Section III presents our novel theoretical analysis of unconstrained SLM methods for EPI ghost correction. Section IV describes new constrained SLM formulations that we propose to overcome the theoretical limitations of unconstrained approaches. Section V presents a systematic evaluation of these approaches with respect to current state-of-the-art approaches. Finally, discussion and conclusions are presented in Sec. VI.

## II. BACKGROUND AND NOTATION

While many SLM descriptions have appeared in the literature, our description of SLM will focus on the perspectives and terminology from the LORAKS framework. For simplicity, we only present a high-level review of LORAKS for the 2D case, and refer interested readers to Refs. [20]–[23] for more general descriptions and additional details.

LORAKS is a flexible constrained reconstruction framework that uses SLM modeling to unify and jointly impose several different classical and widely used MRI reconstruction constraints: limited image support constraints, smooth phase constraints, parallel imaging constraints, and sparsity constraints [20]–[23]. LORAKS is based on the observation that if any of these constraints is appropriate for a given image, then the Nyquist-sampled k-space data for that image will possess shift-invariant linear prediction relationships. These relationships mean that missing/corrupted data samples can be extrapolated or imputed as a weighted linear combination of neighboring points in k-space. Such linear prediction relationships imply that the k-space data will lie in a low-dimensional subspace, and SLM approaches can implicitly learn and impose this subspace structure directly from undersampled/low-quality data.

Importantly, while the data will possess linear prediction relationships and lie in a low-dimensional subspace under assumptions about the image support, phase, etc., the LO-RAKS approach is agnostic to the original source of these relationships. Instead, the approach attempts to identify and utilize all of the linear prediction relationships that are present in the k-space data, regardless of their source. This means that, while LORAKS reconstruction may be easier when support, phase, and parallel imaging constraints are applicable simultaneously, the LORAKS approach can still function when one or more of these constraints is inapplicable, as long as there are sufficient sources of linear predictability in the data.

The basic premise of the LORAKS support constraint [20] is that, if there are large regions of the FOV in which the true image is identically zero and if $s(k_x, k_y)$ represents the Fourier transform of the true image, then there exist infinitely many k-space functions $f(k_x, k_y)$ such that $s(k_x, k_y) * f(k_x, k_y) \approx 0$, where $*$ denotes the standard convolution operation. If we let $\mathbf{k}$ denote the vector of samples of $s(k_x, k_y)$ on the Cartesian Nyquist grid for the FOV and let $\mathbf{f}$ represent the samples of $f(k_x, k_y)$ on the same Cartesian grid, then the convolution relationship can be expressed in matrix-vector form as $\mathcal{P}_C(\mathbf{k})\mathbf{f} \approx 0$, where the operator $\mathcal{P}_C(\mathbf{k})$ forms a Toeplitz-structured convolution matrix (called the LORAKS **C**-matrix [20]) out of the entries of $\mathbf{k}$. Since there are many such vectors $\mathbf{f}$ that satisfy this relationship, we observe that the LORAKS **C**-matrix will be approximately low-rank.

The basic premise of the LORAKS phase constraint [20], [22] is that, if the image has smoothly-varying phase and the image has limited support, then there are infinitely many functions $h(k_x, k_y)$ such that $s(k_x, k_y) * h(k_x, k_y) - \bar{s}(-k_x, -k_y) * \bar{h}(k_x, k_y) \approx 0$, where $\bar{s}(k_x, k_y)$ and $\bar{h}(k_x, k_y)$ are respectively the complex conjugates of $s(k_x, k_y)$ and $h(k_x, k_y)$. Similar to the previous case, this convolution relationship can be expressed in a matrix-vector form as $\mathcal{P}_S(\mathbf{k})\mathbf{h} \approx 0$, where the operator $\mathcal{P}_S(\mathbf{k})$ combines a Toeplitz-structured convolution matrix with a Hankel-structured convolution matrix (resulting in what we call the LORAKS **S**-matrix [20], [22]) out of the entries of $\mathbf{k}$, and $\mathbf{h}$ is the vector of Nyquist samples of $h(k_x, k_y)$ and $\bar{h}(k_x, k_y)$. Since there are many such vectors $\mathbf{h}$ that statisfy this relationship, we observe that the LORAKS **S**-matrix will also be approximately low-rank.

These low-rank matrix constructions are easily generalized to the context of parallel imaging. Specifically, assume that data is acquired from $N_c$ channels, and let $\mathbf{k}_n$ denote the vector of k-space samples from the $n$th channel, and let $\mathbf{k}_{\text{tot}}$ denote the vector containing the k-space samples from all channels. It has been shown that the concatenated matrix

$$\mathbf{C}_P(\mathbf{k}_{\text{tot}}) = \begin{bmatrix} \mathcal{P}_C(\mathbf{k}_1) & \mathcal{P}_C(\mathbf{k}_2) & \cdots & \mathcal{P}_C(\mathbf{k}_{N_c}) \end{bmatrix} \quad (1)$$

will generally have low rank [19], [21], [34], and that the concatenated matrix

$$\mathbf{S}_P(\mathbf{k}_{\text{tot}}) = \begin{bmatrix} \mathcal{P}_S(\mathbf{k}_1) & \mathcal{P}_S(\mathbf{k}_2) & \cdots & \mathcal{P}_S(\mathbf{k}_{N_c}) \end{bmatrix} \quad (2)$$

will also generally have low rank [21]. Note that Eqs. (1) and (2) reduce to the standard single-channel case when $N_c = 1$, so we will use these expressions for both the single-channel and the multi-channel cases.

The observation that these matrices have approximately low rank is valuable, because these low-rank characteristics can be exploited to improve image reconstruction quality. Specifically, by enforcing one or more of these low-rank constraints during image reconstruction, it becomes possible to reconstruct high-quality images from highly accelerated and/or unconventionally sampled k-space data.

The preceding paragraphs described SLM approaches for single-channel and multi-channel image reconstruction for general contexts, but without specialization to ghost correction for EPI. However, as described in the introduction, there is a straightforward analogy between parallel imaging and Nyquist ghost correction. For the sake of simplicity and without loss of generality, we will describe the SLM matrix construction for this case in the context of single-shot imaging with positive and negative readout polarities (denoted RO$^+$ and RO$^-$, respectively), noting that the extension to multi-shot imaging is trivial (obtained by concatenating together the SLM matrices for each shot as if the different shots were coming from different receiver coils in a parallel imaging experiment). Let $\mathbf{k}_{\text{tot}}^+$ and $\mathbf{k}_{\text{tot}}^-$ represent hypothetical vectors of Nyquist-sampled Cartesian k-space data for the two different readout gradient polarities from either a single-channel or multi-channel experiment. Under the assumption that we can treat different readout polarities in the same way we treat different receiver coils in parallel imaging, we expect the matrices

$$\begin{bmatrix} \mathbf{C}_P(\mathbf{k}_{\text{tot}}^+) & \mathbf{C}_P(\mathbf{k}_{\text{tot}}^-) \end{bmatrix} \quad (3)$$

and

$$\begin{bmatrix} \mathbf{S}_P(\mathbf{k}_{\text{tot}}^+) & \mathbf{S}_P(\mathbf{k}_{\text{tot}}^-) \end{bmatrix} \quad (4)$$

to be approximately low-rank.[2] However, due to the form of single-shot EPI imaging, we only measure a subset of the phase encoding lines of $\mathbf{k}_{\text{tot}}^+$ and $\mathbf{k}_{\text{tot}}^-$. Specifically, let the measured data for the RO$^+$ and RO$^-$ be respectively denoted as $\mathbf{d}_{\text{tot}}^+$ and $\mathbf{d}_{\text{tot}}^-$, respectively, with $\mathbf{d}_{\text{tot}}^+ = \mathbf{A}_+ \mathbf{k}_{\text{tot}}^+$ and $\mathbf{d}_{\text{tot}}^- = \mathbf{A}_- \mathbf{k}_{\text{tot}}^-$, where $\mathbf{A}_+$ and $\mathbf{A}_-$ are simple subsampling matrices that extract the measured entries of $\mathbf{k}_{\text{tot}}^+$ and $\mathbf{k}_{\text{tot}}^-$ (i.e., $\mathbf{A}_+$ and $\mathbf{A}_-$ are formed by concatenating the rows of the identity matrix corresponding to the k-space sampling masks for each polarity).

With these definitions, we can now describe previous SLM-based EPI ghost correction methods with a consistent language. For example, the earliest and simplest such approach, ALOHA [28], can be viewed as a special case of the optimization problem

$$\left\{ \hat{\mathbf{k}}_{\text{tot}}^+, \hat{\mathbf{k}}_{\text{tot}}^- \right\} = \underset{\{\mathbf{k}_{\text{tot}}^+, \mathbf{k}_{\text{tot}}^-\}}{\arg\min} J\left( \begin{bmatrix} \mathbf{C}_P(\mathbf{k}_{\text{tot}}^+) & \mathbf{C}_P(\mathbf{k}_{\text{tot}}^-) \end{bmatrix} \right), \quad (5)$$

subject to the additional data-consistency constraints that $\mathbf{A}_+ \hat{\mathbf{k}}_{\text{tot}}^+ = \mathbf{d}_{\text{tot}}^+$ and $\mathbf{A}_- \hat{\mathbf{k}}_{\text{tot}}^- = \mathbf{d}_{\text{tot}}^-$. Here, $J(\cdot)$ is a cost function that depends only on the singular values of its matrix argument, and promotes low-rank solutions. In the sequel, we will use the notation

$$L_{\mathbf{C}}(\mathbf{k}_{\text{tot}}^\pm) = J\left( \begin{bmatrix} \mathbf{C}_P(\mathbf{k}_{\text{tot}}^+) & \mathbf{C}_P(\mathbf{k}_{\text{tot}}^-) \end{bmatrix} \right), \quad (6)$$

---

[2]As noted in the introduction, the widely-used assumption that different readout polarities are different modulations of some original image (which forms the basis for the analogy between parallel imaging and ghost correction) is a simplification of the true imaging physics. However, it should be noted that Eqs. (3) and (4) may still possess low-rank structure even if this assumption is violated. In particular, these LORAKS matrices will have low-rank structure as long as there exist shift-invariant linear prediction relationships in k-space, and the LORAKS approach is agnostic to the original source of such relationships. We believe that deriving the existence of linear prediction relationships in the presence of more realistic models (e.g., accounting for eddy currents) may be feasible, although it is beyond the scope of the present work. Nevertheless, the results we show later with real data seem to imply that linear predictability assumptions are reasonably applicable in the real scenarios we have examined.

where $\mathbf{k}_{\text{tot}}^{\pm}$ concatenates $\mathbf{k}_{\text{tot}}^{+}$ and $\mathbf{k}_{\text{tot}}^{-}$. Similarly, we will also use $L_\mathbf{S}(\cdot)$ to denote the function with the same form as Eq. (6), but switching from the LORAKS $\mathbf{C}$-matrix to the LORAKS $\mathbf{S}$-matrix by replacing all instances of $\mathbf{C}_P$ with $\mathbf{S}_P$.

A popular choice for $J(\cdot)$ in the general low-rank matrix completion literature (and the choice made by Ref. [28]) is the nuclear norm, which is a convex function that is known to encourage minimum-rank solutions [36]. The nuclear norm of a matrix $\mathbf{G}$ is defined as

$$\|\mathbf{G}\|_* = \sum_{i=1}^{\text{rank}(\mathbf{G})} \sigma_i(\mathbf{G}), \quad (7)$$

where $\sigma_i(\mathbf{G})$ is the $i$th singular value of $\mathbf{G}$. Another potential choice of $J(\cdot)$, which was proposed in the original LORAKS work [20] but which has not been used in the EPI ghost correction work by other groups, is defined by

$$J_r(\mathbf{G}) = \sum_{i=r+1}^{\text{rank}(\mathbf{G})} (\sigma_i(\mathbf{G}))^2, \quad (8)$$

where $r$ is a user-selected parameter. This cost function is nonconvex, and $J_r(\mathbf{G})$ will equal zero whenever $\text{rank}(\mathbf{G}) \leq r$. However, if $\text{rank}(\mathbf{G}) > r$, then $J_r(\mathbf{G})$ will be nonzero, and equal to the squared Frobenius norm error that is incurred when $\mathbf{G}$ is optimally approximated by a rank-$r$ matrix. As a result, this cost function will encourage the reconstructed image to have a LORAKS matrix that is approximately rank-$r$ or lower.

The following subsection provides a novel theoretical analysis of the optimization problem from Eq. (5), which reveals that it has several undesirable characteristics.

## III. ANALYSIS OF SLM EPI GHOST CORRECTION

For our analysis, we assume a typical setup in which the nominal fully-sampled k-space dataset has equally-spaced consecutive phase encoding positions. For the sake of brevity, we will assume fully-sampled single-channel EPI imaging[3] in which $\mathbf{d}_{\text{tot}}^{+}$ corresponds to the full set of measured even phase encoding positions, while $\mathbf{d}_{\text{tot}}^{-}$ corresponds to the full set of measured odd phase encoding positions.

Notice that the form of $\mathbf{A}_+$ implies that $\mathbf{A}_+^H \mathbf{A}_+$ is a diagonal projection matrix, and that multiplying any vector of k-space samples by $\mathbf{A}_+^H \mathbf{A}_+$ is equivalent to preserving the values of the even phase encoding lines while setting the values of the odd phase encoding lines to zero. Similarly, $\mathbf{A}_-^H \mathbf{A}_-$ is a diagonal projection matrix, and multiplying any vector of k-space samples by $\mathbf{A}_-^H \mathbf{A}_-$ is equivalent to preserving the values of the odd phase encoding lines while setting the values of the even phase encoding lines to zero. Additionally, we have that $\mathbf{A}_+^H \mathbf{A}_+ = \mathbf{I} - \mathbf{A}_-^H \mathbf{A}_-$, where $\mathbf{I}$ is the identity matrix.

---

[3]Generalized theoretical results for the case of parallel imaging with uniformly undersampled phase encoding can also be derived using the same principles we used for the single-channel fully-sampled case. We have elected not to show these derivations because they are intellectually straightforward extensions of the single-channel fully-sampled case, but require a lot of additional notation to describe.

Using these facts together with the vector space concepts of orthogonal complements and direct sums [37], we know that if $\{\hat{\mathbf{k}}_{\text{tot}}^{+}, \hat{\mathbf{k}}_{\text{tot}}^{-}\}$ obeys the data fidelity constraint from Eq. (5), then there exist corresponding vectors $\mathbf{y}$ and $\mathbf{z}$ such that we can write

$$\begin{aligned} \hat{\mathbf{k}}_{\text{tot}}^{+} &= \mathbf{A}_+^H \mathbf{d}_{\text{tot}}^{+} + \mathbf{A}_-^H \mathbf{A}_- \mathbf{y} \\ \hat{\mathbf{k}}_{\text{tot}}^{-} &= \mathbf{A}_-^H \mathbf{d}_{\text{tot}}^{-} + \mathbf{A}_+^H \mathbf{A}_+ \mathbf{z}. \end{aligned} \quad (9)$$

We have the following theoretical results:

**Theorem 1.** *Given the context described above and arbitrary vectors $\mathbf{y}$ and $\mathbf{z}$, the singular values of the matrix*

$$\begin{bmatrix} \mathbf{C}_P(\hat{\mathbf{k}}_{\text{tot}}^{+}) & \mathbf{C}_P(\hat{\mathbf{k}}_{\text{tot}}^{-}) \end{bmatrix} \quad (10)$$

*are identical to the singular values of the matrix*

$$\begin{bmatrix} \mathbf{C}_P(\tilde{\mathbf{k}}_{\text{tot}}^{+}) & \mathbf{C}_P(\tilde{\mathbf{k}}_{\text{tot}}^{-}) \end{bmatrix}, \quad (11)$$

*with*

$$\begin{aligned} \tilde{\mathbf{k}}_{\text{tot}}^{+} &= \mathbf{A}_+^H \mathbf{d}_{\text{tot}}^{+} - \mathbf{A}_-^H \mathbf{A}_- \mathbf{y} \\ \tilde{\mathbf{k}}_{\text{tot}}^{-} &= \mathbf{A}_-^H \mathbf{d}_{\text{tot}}^{-} - \mathbf{A}_+^H \mathbf{A}_+ \mathbf{z}. \end{aligned} \quad (12)$$

*Note that $\tilde{\mathbf{k}}_{\text{tot}}^{+}$ and $\tilde{\mathbf{k}}_{\text{tot}}^{-}$ from Eq. (12) are identical to the vectors $\hat{\mathbf{k}}_{\text{tot}}^{+}$ and $\hat{\mathbf{k}}_{\text{tot}}^{-}$ appearing in Eq. (9), except that the estimates of the unmeasured data samples have been multiplied by -1.*

The proof of this theorem is sketched in Section S.I of the supplementary material.[4] Some basic intuition for this result is that we can multiply our estimates for the unmeasured k-space lines by -1 without impacting fidelity with the measured data. Due to uniform subsampling of each gradient polarity by a factor of 2, this multiplication procedure is equivalent to applying linear phase in k-space, which corresponds to a spatial shift of the image by half the FOV along the phase encoding dimension (and a $180°$ constant phase offset for the $\text{RO}^-$ polarity). This shifting procedure has no effect on the image support or on the correlations that exist between the different coils, and thus has no impact on the singular values or the rank of the LORAKS matrix. This means that if we have one solution to Eq. (5), then it is easy for us to construct another solution to Eq. (5), and this optimization problem will generally not have a unique useful solution.

The following corollaries formalize some of these statements and provide additional useful insight.

**Corollary 1.** *Equation (5) either has the unique solution $\{\hat{\mathbf{k}}_{\text{tot}}^{+}, \hat{\mathbf{k}}_{\text{tot}}^{-}\} = \{\mathbf{A}_+^H \mathbf{d}_{\text{tot}}^{+}, \mathbf{A}_-^H \mathbf{d}_{\text{tot}}^{-}\}$ which corresponds to zero-filling of the measured data, or it has at least two distinct optimal solutions that share exactly the same cost function value.*

**Corollary 2.** *If the cost function $J(\cdot)$ is chosen to be convex (e.g., the nuclear norm), then the zero-filled solution is always an optimal solution of Eq. (5). If Eq. (5) has more than one optimal solution, then it has infinitely many optimal solutions.*

**Corollary 3.** *Theorem 1 and Corollaries 1 and 2 are still true if we replace $\mathbf{C}_P(\cdot)$ in Eqs. (5), (10), and (11) with $\mathbf{S}_P(\cdot)$.*

Corollary 1 is proven in the supplementary material, and implies that the optimization problem of Eq. (5) either has a trivial undesirable solution corresponding to zero-filling of the measured data, or it is an ill-posed optimization problem that does not possess a unique solution. While some of the solutions to Eq. (5) may be desirable, there are no guarantees that the algorithm we use to minimize Eq. (5) will yield one of these desirable solutions. Corollary 2 is also proven in the supplementary material, and suggests that the use of convex cost functions to impose LORAKS constraints is likely to be suboptimal relative to the use of nonconvex cost functions. It should be noted that, unlike recent Nyquist ghost correction methods [28]–[30] which have made use of the convex nuclear norm, the early structured low-rank matrix completion methods for MRI all made use of nonconvex cost functions [19]–[22]. These nonconvex options are likely to be better for this problem setting. Corollary 3 is stated without proof (but can be proved using an approach that is similar to our proof of Theorem 1), and indicates that the deficiencies of Eq. (5) are not alleviated by switching from the LORAKS $\mathbf{C}$-matrix to the LORAKS $\mathbf{S}$-matrix.

---

[4]This paper has supplementary downloadable material available at http://ieeexplore.ieee.org, provided by the authors.

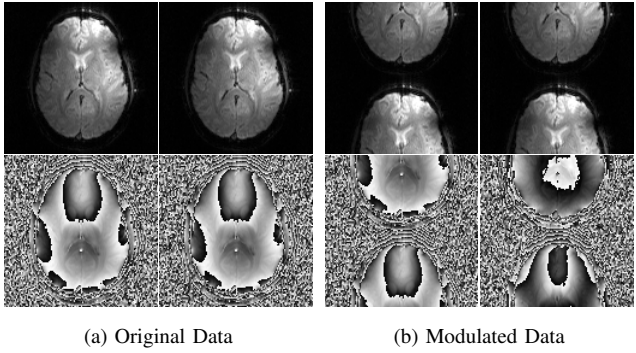

(a) Original Data      (b) Modulated Data

Fig. 1: (a) EPI magnitude (top) and phase (bottom) images corresponding to (left) $\text{RO}^+$ data and (right) $\text{RO}^-$ data. These alias-free images correspond to real data obtained with the PLACE method [11]. (b) Images corresponding to the same data from (a), except that the odd k-space lines for $\text{RO}^+$ and the even k-space lines for $\text{RO}^-$ have been multiplied by -1.

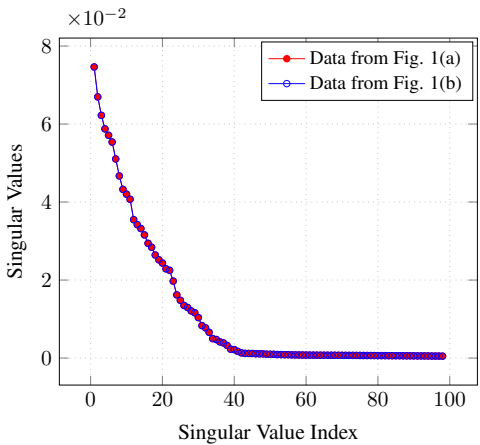

Fig. 2: Plots of the singular values for the LORAKS matrices from Eq. (3) for the k-space datasets from Figs. 1(a) and (b).

Practical illustrations of these theoretical results are shown in Figs. 1–3. Figure 1 shows two different sets of EPI images that are both perfectly consistent with standard fully-sampled EPI data. The difference between the two datasets is the same as the difference between Eqs. (9) and (12). As expected, this k-space phase difference leads to shifting of the images for both $\text{RO}^+$ and $\text{RO}^-$ by half the FOV, as well as adding a constant phase offset for the $\text{RO}^-$ image. Figure 2 shows a plot of the singular values of the LORAKS $\mathbf{C}$-matrices corresponding to these two datasets. As expected from Theorem 1, the singular values are identical in both cases. Figure 3(a,b) illustrates the difference in behavior between convex and nonconvex cost functions $J(\cdot)$. Figure 3(a) shows that the zero-filled solution is a minimum of Eq. (5) in the convex case, as expected from Corollary 2, and that there are many different images with very similar cost function values. Notably, the images from Fig. 1(a) and (b) are not optimal solutions in this convex case, even though they both have high quality and appear to be devoid of ghost artifacts. Figure 3(b) shows that the cost function has more desirable behavior in the nonconvex case (e.g., the zero-filled solution is no longer a minimum of Eq. (5), and there are sharp local minima in the vicinity of the images from Fig. 1), although the solution to Eq. (5) is still not unique in this case as we should expect based on Corollary 1.

While it may be possible to get a useful result from solving Eq. (5), it should be noted that in the presence of multiple global minimizers, it is difficult to ensure that an optimization algorithm will always converge to a desirable minimum. Incorporating additional constraints on the solution is a straightforward way to reduce the ambiguity associated with Eq. (5), and we describe two practical approaches for this in the next section.

## IV. Constrained SLM Formulations

### A. Formulation using SENSE Constraints

A natural approach to imposing additional constraints on SLM reconstruction is to use coil sensitivity map information within the SENSE framework [16], assuming that coil sensitivity profiles are available and that data is acquired using a multi-channel receiver array. This style of approach has been used previously for both EPI reconstruction (assuming ghosts have been precorrected using navigator data) [24] and for navigator-free EPI ghost correction [29], [30]. Our proposed approach can be viewed as a combination of these two previous formulations.

In this work, we propose to use the following formulation for navigator-free EPI ghost correction using SENSE:

$$\{\hat{\boldsymbol{\rho}}^+, \hat{\boldsymbol{\rho}}^-\} = \arg\min_{\{\boldsymbol{\rho}^+,\boldsymbol{\rho}^-\}} \|\mathbf{E}_+\boldsymbol{\rho}^+ - \mathbf{d}_{\text{tot}}^+\|_2^2 \\ + \|\mathbf{E}_-\boldsymbol{\rho}^- - \mathbf{d}_{\text{tot}}^-\|_2^2 + \lambda L_{\mathbf{S}}(\mathbf{k}_{\text{tot}}^{\pm}), \quad (13)$$

subject to the constraints that $\mathbf{E}\boldsymbol{\rho}^+ = \mathbf{k}_{\text{tot}}^+$, that $\mathbf{E}\boldsymbol{\rho}^- = \mathbf{k}_{\text{tot}}^-$, and that $\mathbf{k}_{\text{tot}}^{\pm}$ is the concatenation of $\mathbf{k}_{\text{tot}}^+$ and $\mathbf{k}_{\text{tot}}^-$. In this formulation, we are using SENSE to reconstruct one image for $\text{RO}^+$ ($\boldsymbol{\rho}^+$) and another image for $\text{RO}^-$ ($\boldsymbol{\rho}^-$), and the only coupling that occurs between the two comes from the SLM regularization term. We have also used $\mathbf{E}_+$, $\mathbf{E}_-$, and $\mathbf{E}$ to denote the standard SENSE matrices (mapping from the image domain to k-space) corresponding to RO$^+$ subsampling, RO$^-$ subsampling, and full Nyquist sampling, respectively. These three matrices all use exactly the same sensitivity profiles, and differ only in the associated k-space sampling patterns. In addition, $\lambda$ is a regularization parameter, and we suggest the use of the nonconvex regularizer from Eq. (8) for $L_\mathbf{S}(\cdot)$.

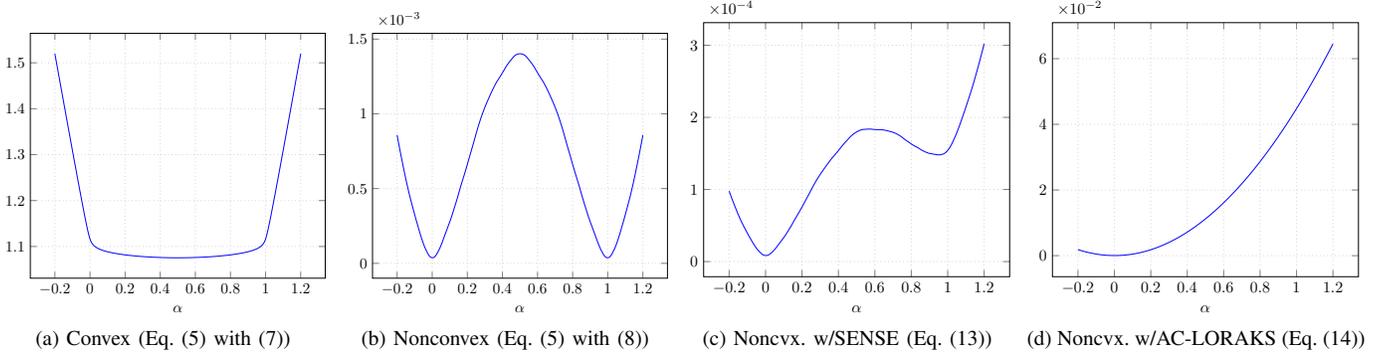

Fig. 3: Letting $\mathbf{k}_1^\pm$ and $\mathbf{k}_2^\pm$ denote the k-space data for the images in Figs. 1(a) and 1(b), respectively, we plot the cost function value $L_\mathbf{C}\left(\alpha\mathbf{k}_1^\pm + (1-\alpha)\mathbf{k}_2^\pm\right)$ as a function of $\alpha$. Setting $\alpha = 0$ yields the cost function value for the images from Fig. 1(a), setting $\alpha = 1$ yields the cost function value for the images from Fig. 1(b), while setting $\alpha = 0.5$ yields the cost function value for the zero-filled solution. Results are shown for different (a) convex and (b-d) nonconvex choices of $J(\cdot)$. The rank parameter $r = 40$ has been used in the nonconvex cases shown in (b-d), both without (a,b) and with (c,d) additional constraints.

A major difference between this proposed approach and our previous SENSE-LORAKS work [24] is the separation of the RO$^+$ and RO$^-$ datasets, which enables navigator-free ghost correction. The main difference between this proposed approach and MUSSELS [29], [30] is that MUSSELS used the LORAKS **C**-matrix and nuclear norm regularization, while we advocate use of the LORAKS **S**-matrix with nonconvex regularization. Another major difference from MUSSELS is that, in the multi-channel case, we use the multi-channel LORAKS matrices (concatenating $2N_c$ SLMs) of Eqs. (3) and (4) [24], instead of concatenating only 2 SLMs formed from the SENSE reconstructions of each gradient polarity.

Equation (13) has been written assuming single-shot data. In the multi-shot case with phase inconsistencies between different shots, we generalize Eq. (13) by reconstructing a separate image for each polarity and each shot, with a separate SENSE encoding matrix and data fidelity term for each. Note that separating the data from different shots increases the effective acceleration factor for each data consistency term and is also associated with additional computational complexity.

There are many ways to solve the optimization problem in Eq. (13). In this paper, we use the simple majorize-minimize approach from Ref. [24], as described in Section S.III of the supplementary material.

As shown in Fig. 3(c), incorporating SENSE constraints changes the shape of the cost function and can help to resolve the uniqueness issues associated with the inverse problem. In particular, we observe that in this case, the extra information provided by SENSE constraints reduces the ambiguity between solutions, and causes the local minimum associated with $\alpha = 0$ to be substantially preferred over the local minimum associated with $\alpha = 1$. However, it appears that in this case, the nonconvex cost function may still have multiple local minima, in which case careful initialization may be necessary to ensure that the iterative approach converges to a good local minimum.

### B. Formulation using AC-LORAKS Constraints

Another natural approach is to use k-space constraints like those of GRAPPA [17], SPIRiT [33], and PRUNO [34]. In this work, we use a formulation based on AC-LORAKS [35] (with strong similarities to PRUNO [34]). Specifically, in the single-shot case, we solve

$$\left\{\hat{\mathbf{k}}_{\text{tot}}^+, \hat{\mathbf{k}}_{\text{tot}}^-\right\} = \arg\min_{\left\{\mathbf{k}_{\text{tot}}^+, \mathbf{k}_{\text{tot}}^-\right\}} \left\|\mathbf{C}_P(\mathbf{k}_{\text{tot}}^+)\mathbf{N}\right\|_F^2 + \left\|\mathbf{C}_P(\mathbf{k}_{\text{tot}}^-)\mathbf{N}\right\|_F^2 + \lambda L_\mathbf{S}(\mathbf{k}_{\text{tot}}^\pm), \quad (14)$$

subject to the constraints that $\mathbf{d}_{\text{tot}}^+ = \mathbf{A}_+\mathbf{k}_{\text{tot}}^+$, $\mathbf{d}_{\text{tot}}^- = \mathbf{A}_-\mathbf{k}_{\text{tot}}^-$, and that $\mathbf{k}_{\text{tot}}^\pm$ is the concatenation of $\mathbf{k}_{\text{tot}}^+$ and $\mathbf{k}_{\text{tot}}^-$. In this expression, $\|\cdot\|_F$ denotes the Frobenius norm, and the matrix $\mathbf{N}$ is an estimate of the approximate right nullspace of a LORAKS **C**-matrix formed from ACS data acquired in a standard parallel imaging calibration pre-scan. The first two terms of Eq. (14) are similar to the first two terms of Eq. (13), in the sense that they impose support and parallel imaging constraints derived from some form of prescan, but do not make any assumptions about the relationship between $\mathbf{k}_{\text{tot}}^+$ and $\mathbf{k}_{\text{tot}}^-$ or the relationship between the image-domain phase characteristics of the calibration data and the image-domain phase characteristics of the EPI data being reconstructed. Similarly, the third terms in Eqs. (13) and (14) are the only terms that introduce coupling between $\mathbf{k}_{\text{tot}}^+$ and $\mathbf{k}_{\text{tot}}^-$, and the only terms that use the LORAKS **S**-matrix to introduce constraints on the image phase. The use of phase constraints is useful both for partial Fourier EPI acquisition and for stabilizing the reconstruction of symmetrically-acquired EPI data [24]. As before, we suggest using the nonconvex regularizer from Eq. (8) for $L_\mathbf{S}(\cdot)$.

As shown in Fig. 3(d), incorporating AC-LORAKS constraints also changes the shape of the cost function. Just like with SENSE constraints, we observe that the additional information provided by the AC-LORAKS constraints leads to less ambiguity, and a clear preference towards the desired solution at $\alpha = 0$. However, the AC-LORAKS approach appears to be even more beneficial than the SENSE approach in this case, since we no longer observe a local minimum associated with $\alpha = 1$ in this case. As a result, we may expect that the AC-LORAKS approach is less sensitive to local minima and yields better solutions than the SENSE approach.

Similar to before, Eq. (14) is written for the single-shot case, but the generalization to multi-shot EPI is straightforward by separating and jointly reconstructing images for each polarity and shot. And similar to Eq. (13), there are also many ways to solve the optimization problem in Eq. (14). These two optimization problems have very similar structure (i.e., the first two terms are least-squares penalties, while the third term encourages low-rank matrix structure), and as a consequence, we have used a minor modification of the algorithm we used for solving Eq. (13) to also solve Eq. (14).

An interesting feature of our proposed AC-LORAKS formulation is that it can potentially work with single-channel data [35]. This is a major advantage over our proposed SENSE formulation, which is not expected to produce good results unless a multiple-channel receiver array is used. It should be noted that while many human MRI experiments use parallel imaging, single-channel reconstruction is still highly relevant in a variety of situations, including animal studies and studies of human anatomy that make use of specialized receiver coil technology (e.g., single-channel prostate coils).

## V. Results

This section describes evaluations of our new LORAKS-based EPI ghost correction methods using navigator-free gradient-echo EPI data acquired from phantoms and *in vivo* human brains. Additional information about the pulse sequence parameters for each dataset is provided in Section S.II of the supplementary material. For each dataset, ACS data (used both for estimating nullspaces and for estimating sensitivity maps) was acquired using the same approach as previously used in DPG [13]. Most of the reconstructions we show in this section estimate separate images for each gradient polarity and each shot, and some also estimate separate images for each coil. While various approaches exist for combining together multiple images from different coils/polarities/shots for visualization, for simplicity and consistency we have combined the multiple images into a single image using principal component analysis, which is a standard method for parallel imaging coil compression/combination [38], [39]. Unless otherwise specified, our LORAKS-based results also always use the nonconvex regularization penalty from Eq. (8), with the rank threshold $r$ chosen based on the singular values of the LORAKS matrix formed from ACS data. Specifically, $r$ was chosen as the point at which the plot of the singular values appears to flatten out, which is a standard approach to matrix rank estimation in the presence of noise. For methods that use regularization parameters, $\lambda$ was initially set to a small value ($\lambda = 10^{-3}$), and if necessary based on visual assessment of image ghost artifacts, was gradually increased until good reconstructions were observed. LORAKS-based reconstructions were performed based on adaptations of publicly-available code [40].

Figure 4 shows a comparison of different parallel imaging reconstruction and EPI ghost correction methods for *in vivo* single-shot EPI data. A gold standard image with fully-sampled RO$^+$ data and fully-sampled RO$^-$ images was obtained using PLACE [11] with a 32-channel receiver coil and an $128\times128$ acquisition matrix. We also acquired standard fully-sampled EPI (acceleration factor $R = 1$, with each gradient polarity undersampled by a factor of two) and prospectively accelerated EPI acquisitions for a range of acceleration factors ($R = 2, 3, 4$). Additionally, the acceleration factor of $R = 5$ was simulated by retrospectively under-sampling the PLACE data. Reconstructions were performed using unconstrained LORAKS as in Eq. (5) but using the LORAKS **S**-matrix, LORAKS with SENSE constraints as in Eq. (13) in both convex (Eq. (7)) and nonconvex (Eq. (8)) variations, and LORAKS with AC-LORAKS constraints as in Eq. (14). For SENSE-based reconstruction, sensitivity profiles were estimated using ESPIRiT [41]. For comparison, we also performed MUSSELS reconstruction [29], [30] and independent SENSE reconstruction of each gradient polarity [12] without LORAKS-based regularization (equivalent to setting $\lambda = 0$ in Eq. (13)).

The figure shows that SENSE without SLM regularization works well for low-acceleration factors, though faces challenges at high acceleration factors. This behavior is expected as an EPI acceleration factor of $R = 5$ is an effective acceleration factor of $R = 10$ for each readout polarity, which is a very challenging case for SENSE reconstruction. We also observe that unconstrained LORAKS reconstruction has severe problems, as should be expected based on our theoretical analysis of Eq. (5). The results also show that the convex SLM approaches (MUSSELS and LORAKS with SENSE and convex regularization) are effective at low acceleration factors, but start demonstrating artifacts as the acceleration factor increases. On the other hand, both of our proposed new formulations are substantially more successful, achieving high quality reconstruction results even at very high acceleration factors. This result is consistent with our theoretical expectations from Corollaries 1-3 that nonconvex cost functions can lead to a better-posed reconstruction problem. At the highest acceleration factors, LORAKS with AC-LORAKS constraints was more effective than LORAKS with SENSE constraints, which displayed unresolved aliasing artifacts.

The data shown in Fig. 4 was also reconstructed using the state-of-the-art DPG method [13] using the same ACS data, and a comparison against LORAKS with AC-LORAKS constraints is shown in Fig. 5. While DPG generally works well, a close examination of the reconstructed magnitude and phase images demonstrates that DPG still has small residual ghost artifacts that are not present in the LORAKS-based reconstruction. These artifacts are particularly visible in the phase images, since the phase is highly sensitive to ghosting

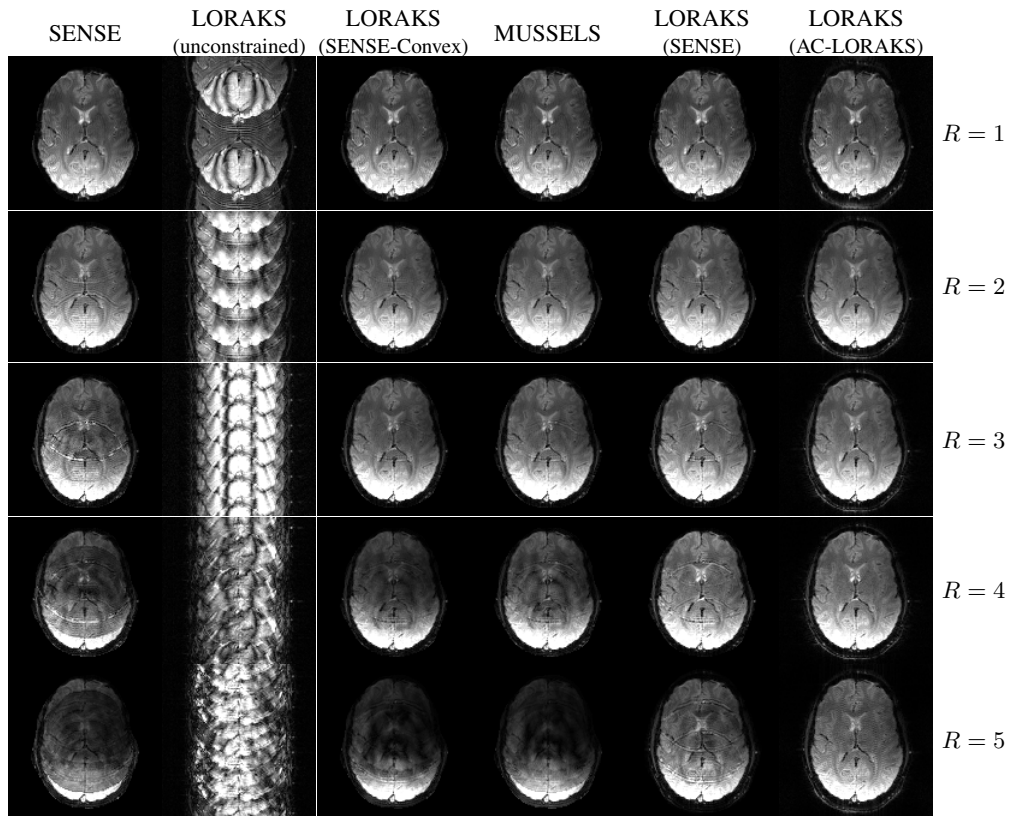

Fig. 4: Comparison of different reconstruction techniques using prospectively undersampled (except the $R=5$ case, which is retrospectively undersampled) *in vivo* single-shot EPI data at different parallel imaging acceleration factors.

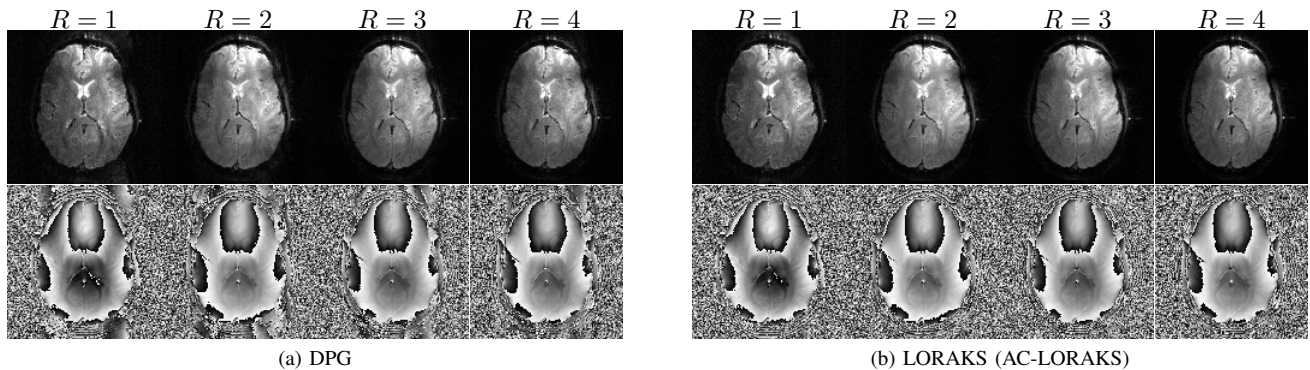

(a) DPG

(b) LORAKS (AC-LORAKS)

Fig. 5: Comparison between (a) DPG and (b) LORAKS with AC-LORAKS constraints for the real single-shot EPI *in vivo* brain data from Fig. 4. Instead of showing coil-combined images, a single representative channel is shown to avoid contamination of the phase characteristics induced by coil combination.

in regions of the image where the magnitude is small. A deeper examination of the data leads us to believe that the ghost artifacts we see for DPG are the result of systematic changes (between the relative phases of the different gradient polarities) that have occurred in part due to the length of time that passed between the collection of the ACS data and the acquisition of the accelerated EPI data that is being reconstructed.

The data used in Figs. 4 and 5 was prospectively sampled, but it is hard to quantify accuracy for prospectively sampled data because the image phase characteristics (due to eddy curents, etc.) can vary as a function of sequence parameters like the acceleration factor, and because our subject is living and breathing (which leads to variations over time). An illustration of the phase differences between $RO^+$ and $RO^-$ for different acceleration factors is shown in supplementary Fig. S1. To enable a quantitative comparison of different approaches, we also performed reconstructions of retrospectively undersampled versions of the fully sampled PLACE data. Results are shown in supplementary Fig. S2, with normalized root-mean-squared error (NRMSE) shown in supplementary Table S1. The retrospective results are consistent with our prospective results, and the quantitative NRMSE results are consistent with our qualitative evaluations.

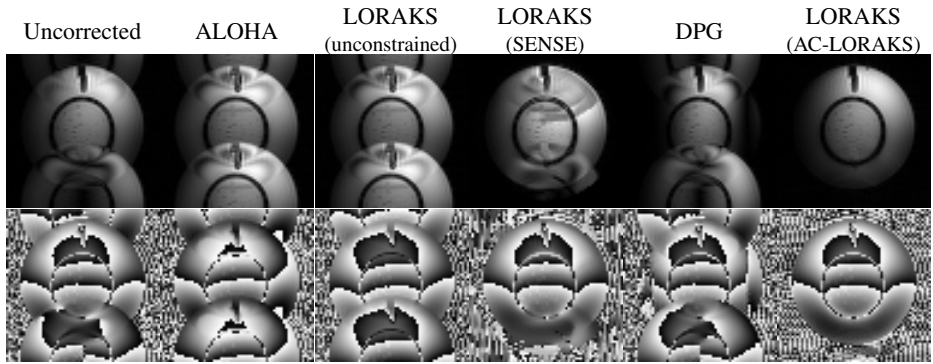

Fig. 6: (top) Magnitude and (bottom) phase images corresponding to reconstruction of unaccelerated ($R=1$) single-channel single-shot EPI phantom data.

The data shown in Figs. 4 and 5 was acquired with an axial slice orientation, and therefore had less phase mismatch between $RO^+$ and $RO^-$ than it could have if we had used a less-conventional slice orientation. In addition, the slice we used was far from sources of field inhomogeneity, and therefore had a relatively smooth phase profile. To demonstrate a more complicated case, we have also performed ghost correction of EPI data acquired with a double-oblique slice orientation. Double-oblique orientation is non-traditional for EPI imaging of the brain, but is known to give rise to 2D nonlinear phase-differences between gradient polarities due, e.g., to concomitant field effects [8], [13], [42]–[44]. In addition, the double-oblique slice we selected passes close to air-tissue interfaces and has a substantially less-smooth phase profile. We have performed reconstructions of prospectively undersampled double-oblique data (similar to Figs. 4 and 5), and the results are shown in supplementary Figs. S3-S5. We have also performed reconstructions of retrospectively under-sampled double-oblique data (similar to supplementary Fig. S2 and supplementary Table S1), and the results are shown in supplementary Fig. S6 and supplementary Table S2. As can be seen from the phase difference maps in supplementary Fig. S4, we have more significant nonlinear 2D phase mismatches between $RO^+$ and $RO^-$ in this case. Our results with double-oblique data are consistent with our results from axial data, suggesting that our proposed approaches can still work well in more challenging scenarios.

The results described above were all based on data acquired with a somewhat loose FOV, which may be beneficial for the use of support constraints. However, there are also imaging scenarios of interest in which the FOV is much tighter. To test performance in the presence of a tight FOV, we acquired similar data to that shown in Figs. 4 and 5, but with half the FOV along the phase encoding dimension. This causes our gold standard image to demonstrate aliasing. To make the reconstruction problem even more challenging, we also simulated an additional 2D nonlinear phase difference between $RO^+$ and $RO^-$, and this phase difference was chosen differently for the ACS data and the gold standard used for simulation. Since it is difficult to apply conventional SENSE-based methods in the presence of a tight FOV (because of the presence of aliasing in the gold standard), we only performed reconstructions using methods that do not use sensitivity maps, i.e., DPG and LORAKS with AC-LORAKS constraints. The results are reported in supplementary Fig. S7. As can be seen, DPG does not perform very well in this very complicated scenario, particularly because of the phase mismatch between the ACS data and the data being reconstructed. On the other hand, AC-LORAKS is substantially more successful.

In addition to navigator-free multi-channel settings, the proposed methods were also evaluated in navigator-free single-channel settings, which are expected to be substantially more challenging. Single-channel datasets were obtained by isolating the information from a single coil in multi-channel acquisitions. Reconstructions were performed using ALOHA [28] (Eq. (5) with Eq. (7)), unconstrained LORAKS (Eq. (5) with Eq. (8) but using the LORAKS $\mathbf{S}$-matrix), LORAKS with SENSE constraints (Eq. (13)), DPG [13], and LORAKS with AC-LORAKS constraints (Eq. (14)). Note that, since sensitivity-map estimation is not feasible in the single-channel setting, our SENSE-based results used a binary support mask (that has value 1 inside the support of the image and value 0 everywhere else) in place of a coil sensitivity map. This has the effect of imposing prior knowledge of the image support on the reconstructed image. Note also that DPG was not originally designed to be used with single-channel data, although the formulation can still be applied to the single-channel case. Unaccelerated ($R=1$) single-channel single-shot EPI results are shown for a phantom dataset ($64\times64$ acquisition matrix) in Fig. 6 and for one channel of the previous *in vivo* human brain dataset in Fig. 7. The results are consistent in both cases. Images obtained without compensating the mismatch between $RO^+$ and $RO^-$ have obvious ghost artifacts, and these artifacts are not solved (and are potentially even amplified) when using unconstrained SLM approaches (i.e., ALOHA or LORAKS without constraints). The LORAKS reconstruction with "SENSE" constraints (i.e., support constraints) helps to eliminate some of the ghost artifacts that appeared outside the support of the original object, although residual aliasing artifacts are still observed within the support of the object. These artifacts are most visible in the phantom image, though close inspection also reveals the appearance of aliasing artifacts in the brain image. We also observe that DPG is unsuccessful in this single-channel case, which we believe is due both to the difficulties of the single-channel problem as well as systematic differences between the ACS data and the data being reconstructed. On the other hand, the LORAKS results with AC-LORAKS constraints are substantially more successful than any of the previous methods.

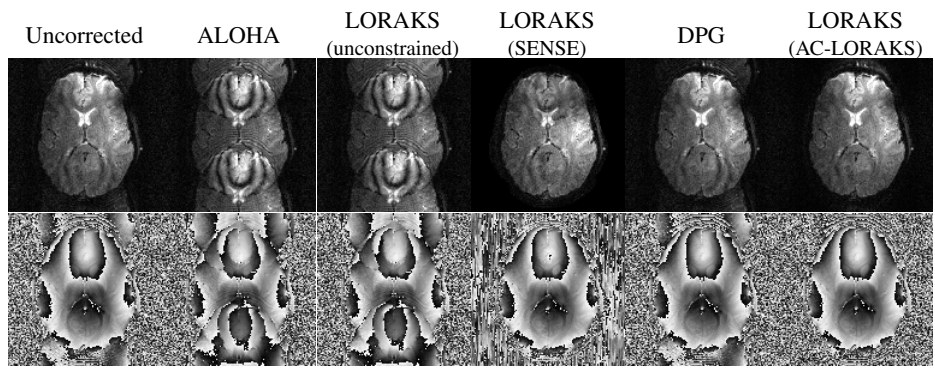

Fig. 7: (top) Magnitude and (bottom) phase images corresponding to reconstruction of unaccelerated ($R=1$) single-channel single-shot EPI *in vivo* human brain data.

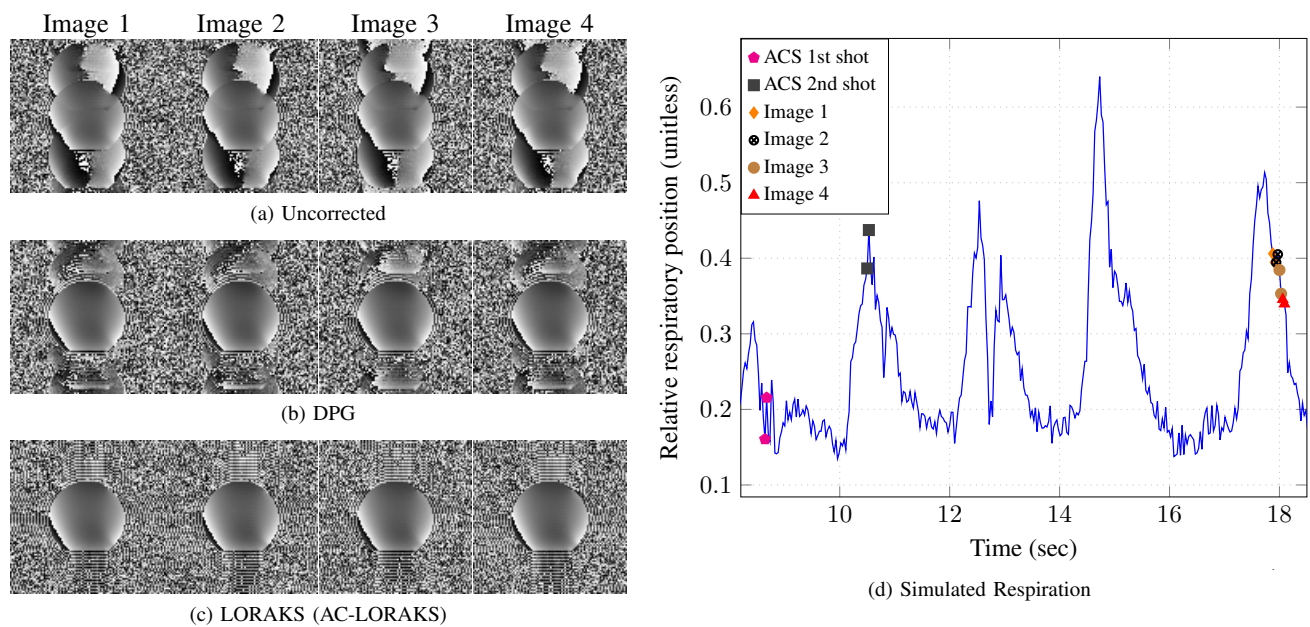

Fig. 8: Evaluation with multi-shot EPI data for a phantom with simulated respiratory effects. A representative set of four segmented images is shown, extracted from a longer acquisition spanning several minutes. (a) Phase images corresponding to one channel of the data, with images reconstructed without compensating for the mismatches between different shots and different gradient polarities. (b) Phase images for one channel of the DPG reconstruction, with reconstruction performed using multi-channel data. (c) Phase images for single-channel LORAKS reconstruction with AC-LORAKS constraints. (d) Plot showing the relative respiratory position across EPI shots (TR=60msec), as measured with an ultrasound transducer coupled to a respiratory phantom [18]. The line-plot peaks show points where the phantom air bag is maximally inflated. The sampling times for the ACS data and for the shots used to generate each of the two-shot images from (a)-(c) are marked as labeled in the legend.

We also applied the different reconstruction approaches to single-channel double-oblique data, for the same reasons as in the multi-channel case. Results are shown in supplementary Fig. S8, and are consistent with the results observed with axial data. One potential criticism of the single-channel results we have shown is that, at least for these datasets, the phase mismatch between RO$^+$ and RO$^-$ is not very severe, which causes the uncorrected reconstructions to have relatively low levels of ghost artifact. To demonstrate the performance in a more severe case, we performed reconstruction of this dataset again with an exaggerated phase mismatch between RO$^+$ and RO$^-$. The results of this simulation are shown in supplementary Fig. S9. As can be seen, this case demonstrates much stronger ghost artifacts without correction. However, similar to the previous cases, LORAKS with AC-LORAKS constraints is still the most successful at reducing ghost artifacts.

Reconstructions were also performed using multi-shot data. Figure 8 shows results using fully sampled $R=1$ two-

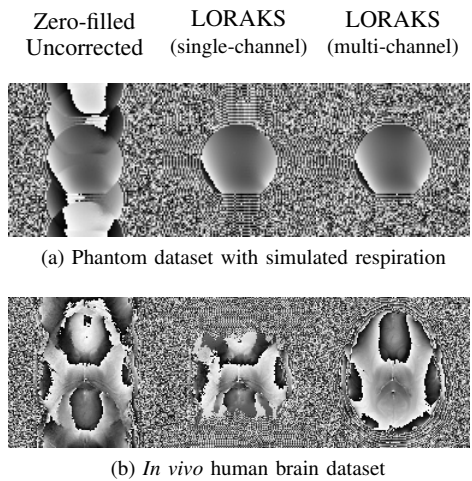

(a) Phantom dataset with simulated respiration

(b) *In vivo* human brain dataset

Fig. 9: Images reconstructed using LORAKS with AC-LORAKS constraints for accelerated ($R = 2$) data in both single-channel and multi-channel contexts. (a) Phantom dataset with simulated respiration. (b) *In vivo* human brain dataset. Phase images from a single channel are shown for zero-filled data reconstructed without compensating for mismatches between the gradient polarities, LORAKS reconstruction with AC-LORAKS constraints from single-channel data, and LORAKS reconstruction with AC-LORAKS constraints from multi-channel data.

shot data (128×128 acquisition matrix, 12-channel receiver coil) at different time points for a phantom with physically-simulated respiration effects [18]. Note that with $R = 1$ and two-shots, each gradient polarity for each shot has an effective undersampling factor of 4. LORAKS with AC-LORAKS constraints is compared against DPG for segmented EPI [45], and to make reconstruction even more challenging for LORAKS, LORAKS reconstruction was performed from single-channel data while DPG was provided with the full set of multi-channel data. Due to simulated respiration, there are mismatches between the measured ACS data and the EPI data being reconstructed. DPG is not robust to these mismatches, and displays residual ghost artifacts with time-varying characteristics. On the other hand, LORAKS with AC-LORAKS demonstrates robustness against the time-varying changes in this dataset, even despite the challenging single-channel multi-shot nature of this reconstruction problem.

For further insight, Fig. 9(a) shows an even more challenging case where LORAKS with AC-LORAKS constraints is used to reconstruct a single-shot of the previous multi-shot dataset. Note that this corresponds to EPI with an acceleration factor of $R = 2$ (i.e., an effective acceleration factor of $R = 4$ for each gradient polarity). Remarkably, we observe that single-channel LORAKS with AC-LORAKS constraints is still successful in this very difficult scenario, with similar quality to that obtained using multi-channel LORAKS with AC-LORAKS constraints. However, it is also important for us to point out that single-channel reconstruction with $R = 2$ is not always successful, as illustrated in Fig. 9(b) using previously described *in vivo* human brain data. In this brain case, we observe that single-channel LORAKS reconstruction is unsuccessful at correctly reconstructing the image, while the multi-channel case (also shown in Fig. 5) yields accurate results as described previously. We suspect that the difference between the phantom result and the *in vivo* result is explained by differences in the size of the FOV relative to the size of the object. Specifically, the size of the FOV is more than twice the size of the phantom, while the same is not true for the *in vivo* data. We expect that LORAKS-based single-channel image reconstruction will be harder for tight FOVs than it is for larger FOVs that contain a larger amount of empty space.

## VI. DISCUSSION AND CONCLUSIONS

This paper derived novel theoretical results for EPI ghost correction based on structured low-rank matrix completion approaches. Key theoretical results include the observation that the corresponding matrix completion problem is ill-posed in the absence of additional constraints, and that convex formulations have undesirable characteristics that are somewhat mitigated by the use of nonconvex formulations. These theoretical results led to two novel problem formulations that use additional constraints and nonconvex regularization to avoid the problems associated with ill-posedness. Our results showed that these new approaches are both effective relative to state-of-the-art ghost correction techniques like DPG, and are capable of handling nonlinear 2D phase mismatches between RO$^+$ and RO$^-$. It was also observed that the proposed variation that uses AC-LORAKS constraints appears to be more effective than the proposed variation that uses SENSE constraints in challenging scenarios with single-channel data or highly-accelerated multi-channel data. In addition, we were surprised to observe that the variation using AC-LORAKS constraints can even be successful when applied to undersampled single-channel data. We believe that these approaches will prove valuable across a range of different applications, but especially those in which navigator-based ghost correction methods are ineffective, in which the need for navigation places undesirable constraints on the minimum achievable echo time or repetition time, in cases where single-channel imaging is unavoidable, or in cases where imaging conditions are likely to change over time over the duration of a long experiment (e.g., functional imaging or diffusion imaging).

This paper focused on simple proof-of-principle demonstrations of the proposed approaches, and there are opportunities for a variety of improvements. For example, the choices of $r$ and $\lambda$ were made manually using heuristic approaches in this paper, though its likely that these same decision processes could be made automatically using standard techniques for rank estimation (e.g., [46]) or ghost correction parameter tuning (e.g., [7]). Additionally, extensions to the case of simultaneous multi-slice imaging are possible within the LORAKS framework [47], and are likely to be practically useful in a range of experiments given the advantages and the modern popularity of combining this approach with EPI [48].

While we derived our new theory and methods in the context of EPI ghost correction, we believe that this paper also has more general consequences. For example, we believe that it is straightforward to generalize our theoretical results to show that unconstrained LORAKS-based reconstruction will be ill-posed for any application that uses uniformly-undersampled Cartesian k-space trajectories (also see similar comments in Ref. [24]). We also believe that the combination of LORAKS with additional constraints will always be beneficial when the constraints are accurate, and so encourage the use of additional constraints whenever the LORAKS reconstruction problem is ill-posed. In addition, we believe that our novel LORAKS formulation with AC-LORAKS constraints is an important innovation that is likely to be useful in other applications, similar to how LORAKS with SENSE constraints has already proven to be useful in other settings [24], [49]. For example, it may be beneficial to use AC-LORAKS constraints to augment existing SLM approaches that have recently been proposed for gradient delay estimation in non-Cartesian MRI [50], [51]. Finally, while recent work has proposed the use of convex LORAKS-based formulations, our empirical experience since we first started exploring LORAKS several years ago [20], [52] has consistently been that nonconvex formulations are substantially more powerful than convex ones.

## References

[1] P. Mansfield, "Multi-planar image formation using NMR spin echoes," *J. Phys. C: Solid State Phys.*, vol. 10, p. L55, 1977.
[2] M. A. Bernstein, K. F. King, and X. J. Zhou, *Handbook of MRI Pulse Sequences*. Burlington: Elsevier Academic Press, 2004.
[3] H. Bruder, H. Fischer, H.-E. Reinfelder, and F. Schmitt, "Image reconstruction for echo planar imaging with nonequidistant k-space sampling," *Magn. Reson. Med.*, vol. 23, pp. 311–323, 1992.
[4] X. Hu and T. H. Le, "Artifact reduction in EPI with phase-encoded reference scan," *Magn. Reson. Med.*, vol. 36, pp. 166–171, 1996.
[5] N.-k. Chen and A. M. Wyrwicz, "Removal of EPI Nyquist ghost artifacts with two-dimensional phase correction," *Magn. Reson. Med.*, vol. 51, pp. 1247–1253, 2004.
[6] M. H. Buonocore and L. Gao, "Ghost artifact reduction for echo planar imaging using image phase correction," *Magn. Reson. Med.*, vol. 38, pp. 89–100, 1997.
[7] S. Skare, D. Clayton, R. Newbould, M. Moseley, and R. Bammer, "A fast and robust minimum entropy based non-interactive Nyquist ghost correction algorithm," in *Proc. Int. Soc. Magn. Reson. Med.*, 2006, p. 2349.
[8] D. Xu, K. F. King, Y. Zur, and R. S. Hinks, "Robust 2D phase correction for echo planar imaging under a tight field-of-view," *Magn. Reson. Med.*, vol. 64, pp. 1800–1813, 2010.
[9] U. Yarach, M.-H. In, I. Chatnuntawech, B. Bilgic, F. Godenschweger, H. Mattern, A. Sciarra, and O. Speck, "Model-based iterative reconstruction for single-shot EPI at 7T," *Magn. Reson. Med.*, 2017, Early View.
[10] J. D. Ianni, E. B. Welch, and W. A. Grissom, "Ghost reduction in echo-planar imaging by joint reconstruction of images and line-to-line delays and phase errors," *Magn. Reson. Med.*, 2017, Early View.
[11] Q.-S. Xiang and F. Q. Ye, "Correction for geometric distortion and N/2 ghosting in EPI by phase labeling for additional coordinate encoding (PLACE)," *Magn. Reson. Med.*, vol. 57, pp. 731–741, 2007.
[12] W. S. Hoge, H. Tan, and R. A. Kraft, "Robust EPI Nyquist ghost elimination via spatial and temporal encoding," *Magn. Reson. Med.*, vol. 64, pp. 1781–1791, 2010.
[13] W. S. Hoge and J. R. Polimeni, "Dual-polarity GRAPPA for simultaneous reconstruction and ghost correction of echo planar imaging data," *Magn. Reson. Med.*, vol. 76, pp. 32–44, 2016.
[14] H.-C. Chang and N.-K. Chen, "Joint correction of Nyquist artifact and minuscule motion-induced aliasing artifact in interleaved diffusion weighted EPI data using a composite two-dimensional phase correction procedure," *Magn. Reson. Imag.*, vol. 34, pp. 974–979, 2016.
[15] V. B. Xie, M. Lyu, Y. Liu, Y. Feng, and E. X. Wu, "Robust EPI Nyquist ghost removal by incorporating phase error correction with sensitivity encoding (PEC-SENSE)," *Magn. Reson. Med.*, 2017, Early View.
[16] K. P. Pruessmann, M. Weiger, M. B. Scheidegger, and P. Boesiger, "SENSE: Sensitivity encoding for fast MRI," *Magn. Reson. Med.*, vol. 42, pp. 952–962, 1999.
[17] M. A. Griswold, P. M. Jakob, R. M. Heidemann, M. Nittka, V. Jellus, J. Wang, B. Kiefer, and A. Haase, "Generalized autocalibrating partially parallel acquisitions (GRAPPA)," *Magn. Reson. Med.*, vol. 47, pp. 1202–1210, 2002.
[18] W. S. Hoge, F. R. Preiswerk, J. R. Polimeni, S. Yengul, P. A. Ciris, and B. Madore, "Ultrasound monitoring of a respiratory phantom for the development and validation of segmented EPI reconstruction methods," in *Proc. Int. Soc. Magn. Reson. Med.*, 2017, p. 1307.
[19] P. J. Shin, P. E. Z. Larson, M. A. Ohliger, M. Elad, J. M. Pauly, D. B. Vigneron, and M. Lustig, "Calibrationless parallel imaging reconstruction based on structured low-rank matrix completion," *Magn. Reson. Med.*, vol. 72, pp. 959–970, 2014.
[20] J. P. Haldar, "Low-Rank Modeling of Local-Space Neighborhoods (LORAKS) for Constrained MRI," *IEEE Trans. Med. Imag.*, vol. 33, pp. 668–681, 2014.
[21] J. P. Haldar and J. Zhuo, "P-LORAKS: Low-rank modeling of local k-space neighborhoods with parallel imaging data," *Magn. Reson. Med.*, vol. 75, pp. 1499–1514, 2015.
[22] J. P. Haldar, "Low-rank modeling of local k-space neighborhoods: From phase and support constraints to structured sparsity," in *Wavelets and Sparsity XVI, Proc. SPIE 9597*, 2015, p. 959710.
[23] J. P. Haldar and T. H. Kim, "Computational imaging with LORAKS: Reconstructing linearly predictable signals using low-rank matrix regularization." in *Proc. Asilomar Conf. Sig. Sys. Comp.*, 2017.
[24] T. H. Kim, K. Setsompop, and J. P. Haldar, "LORAKS makes better SENSE: Phase-constrained partial Fourier SENSE reconstruction without phase calibration," *Magn. Reson. Med.*, vol. 77, pp. 1021–1035, 2016.
[25] K. H. Jin, D. Lee, and J. C. Ye, "A general framework for compressed sensing and parallel MRI using annihilating filter based low-rank Hankel matrix," *IEEE Trans. Comput. Imaging*, vol. 2, pp. 480–495, 2016.
[26] G. Ongie and M. Jacob, "Off-the-grid recovery of piecewise constant images from few Fourier samples," *SIAM J. Imaging Sci.*, vol. 9, pp. 1004–1041, 2016.
[27] P. Kellman and E. R. McVeigh, "Phased array ghost elimination," *NMR Biomed.*, vol. 19, pp. 352–361, 2006.
[28] J. Lee, K. H. Jin, and J. C. Ye, "Reference-free EPI Nyquist ghost correction using annihilating filter-based low rank Hankel matrix for K-space interpolation," *Magn. Reson. Med.*, vol. 76, pp. 1775–1789, 2016.
[29] M. Mani, M. Jacob, D. Kelley, and V. Magnotta, "Multi-shot sensitivity-encoded diffusion data recovery using structured low-rank matrix completion (MUSSELS)," *Magn. Reson. Med.*, vol. 78, pp. 494–507, 2017.
[30] M. Mani, V. Magnotta, D. Kelley, and M. Jacob, "Comprehensive reconstruction of multi-shot multi-channel diffusion data using MUSSELS," in *Proc. IEEE Eng. Med. Bio. Conf.*, 2016, pp. 1107–1110.
[31] M. Lyu, M. Barth, V. B. Xie, Y. Liu, Y. Feng, and E. X. Wu, "Robust 2D Nyquist ghost correction for simultaneous multislice (SMS) EPI using phase error correction SENSE and virtual coil SAKE," in *Proc. Int. Soc. Magn. Reson. Med.*, 2017, p. 515.
[32] R. A. Lobos, T. H. Kim, W. S. Hoge, and J. P. Haldar, "Navigator-free EPI ghost correction using low-rank matrix modeling: Theoretical insights and practical improvements," in *Proc. Int. Soc. Magn. Reson. Med.*, 2017, p. 0449.
[33] M. Lustig and J. M. Pauly, "SPIRiT: Iterative self-consistent parallel imaging reconstruction from arbitary *k*-space," *Magn. Reson. Med.*, vol. 65, pp. 457–471, 2010.
[34] J. Zhang, C. Liu, and M. E. Moseley, "Parallel reconstruction using null operations," *Magn. Reson. Med.*, vol. 66, pp. 1241–1253, 2011.
[35] J. P. Haldar, "Autocalibrated LORAKS for fast constrained MRI reconstruction," in *Proc. IEEE Int. Symp. Biomed. Imag.*, 2015, pp. 910–913.
[36] B. Recht, M. Fazel, and P. A. Parrilo, "Guaranteed minimum-rank solutions of linear matrix equations via nuclear norm minimization," *SIAM Rev.*, vol. 52, pp. 471–501, 2010.
[37] D. G. Luenberger, *Optimization by Vector Space Methods*. New York: John Wiley & Sons, 1969.
[38] M. Buehrer, K. P. Pruessmann, P. Boesiger, and S. Kozerke, "Array compression for MRI with large coil arrays," *Magn. Reson. Med.*, vol. 57, pp. 1131–1139, 2007.
[39] F. Huang, S. Vijayakumar, Y. Li, S. Hertel, and G. R. Duensing, "A software channel compression technique for faster reconstruction with many channels," *Magn. Reson. Imag.*, vol. 26, pp. 133–141, 2008.
[40] J. P. Haldar, "Low-rank modeling of local k-space neighborhoods (LORAKS): Implementation and examples for reproducible research,"

University of Southern California, Los Angeles, CA, Tech. Rep. USC-SIPI-414, April 2014, http://sipi.usc.edu/reports/abstracts.php?rid=sipi-414.

[41] M. Uecker, P. Lai, M. J. Murphy, P. Virtue, M. Elad, J. M. Pauly, S. S. Vasanawala, and M. Lustig, "ESPIRiT – an eigenvalue approach to autocalibrating parallel MRI: Where SENSE meets GRAPPA," *Magn. Reson. Med.*, vol. 71, pp. 990–1001, 2014.

[42] S. B. Reeder, E. Atalar, A. Z. Faranesh, and E. R. McVeigh, "Referenceless interleaved echo-planar imaging," *Magn. Reson. Med.*, vol. 41, pp. 87–94, 1999.

[43] S. M. Grieve, A. M. Blamire, and P. Styles, "Elimination of Nyquist ghosting caused by read-out to phase-encode gradient cross-terms in EPI," *Magn. Reson. Med.*, vol. 47, pp. 337–343, 2002.

[44] N.-K. Chen, A. V. Avram, and A. W. Song, "Two-dimensional phase cycled reconstruction for inherent correction of echo-planar imaging Nyquist artifacts," *Magn. Reson. Med.*, vol. 66, pp. 1057–1066, 2011.

[45] W. S. Hoge and J. R. Polimeni, "Artifact correction in accelerated-segmented EPI data via dual-polarity GRAPPA," in *Proc. Int. Soc. Magn. Reson. Med.*, 2017, p. 449.

[46] E. J. Candès, C. A. Sing-Long, and J. D. Trzasko, "Unbiased risk estimates for singular value thresholding and spectral estimators," *IEEE Trans. Signal Process.*, vol. 61, pp. 4643–4657, 2013.

[47] T. H. Kim and J. P. Haldar, "SMS-LORAKS: Calibrationless simultaneous multislice MRI using low-rank matrix modeling," in *Proc. IEEE Int. Symp. Biomed. Imag.*, 2015, pp. 323–326.

[48] B. A. Poser and K. Setsompop, "Pulse sequences and parallel imaging for high spatiotemporal resolution MRI at ultra-high field," *NeuroImage*, 2017, Early View.

[49] T. H. Kim, B. Bilgic, D. Polak, K. Setsompop, and J. P. Haldar, "Wave-LORAKS for faster wave-CAIPI MRI," in *Proc. Int. Soc. Magn. Reson. Med.*, 2017, p. 1037.

[50] W. Jiang, P. E. Z. Larson, and M. Lustig, "Simultaneous estimation of auto-calibration data and gradient delays in non-Cartesian parallel MRI using low-rank constraints," in *Proc. Int. Soc. Magn. Reson. Med.*, 2016, p. 939.

[51] M. Mani, S. Poddar, V. Magnotta, and M. Jacob, "Trajectory correction of radial data using MUSSELS," in *Proc. Int. Soc. Magn. Reson. Med.*, 2017, p. 1197.

[52] J. P. Haldar, "Calibrationless partial Fourier reconstruction of MR images with slowly-varying phase: A rank-deficient matrix recovery approach," in *ISMRM Workshop on Data Sampling & Image Reconstruction*, Sedona, 2013.

# Supplementary Material for "Navigator-free EPI Ghost Correction with Structured Low-Rank Matrix Models: New Theory and Methods"

Rodrigo A. Lobos, *Student Member, IEEE*, Tae Hyung Kim, *Student Member, IEEE*,
W. Scott Hoge, *Member, IEEE*, and Justin P. Haldar, *Senior Member, IEEE*

## S.I. Proofs

*Proof of Theorem 1*

Assuming the notation of Eq. (5), let $\mathbf{B}$ denote the zero-filled LORAKS matrix

$$\mathbf{B} = \begin{bmatrix} \mathbf{C}_P(\mathbf{A}_+^H \mathbf{d}_{\text{tot}}^+) & \mathbf{C}_P(\mathbf{A}_-^H \mathbf{d}_{\text{tot}}^-) \end{bmatrix} \quad (\text{S.1})$$

and let $\mathbf{D}$ denote the matrix corresponding to the unmeasured data samples

$$\mathbf{D} = \begin{bmatrix} \mathbf{C}_P(\mathbf{A}_-^H \mathbf{A}_- \mathbf{y}) & \mathbf{C}_P(\mathbf{A}_+^H \mathbf{A}_+ \mathbf{z}) \end{bmatrix}. \quad (\text{S.2})$$

Due to the way the LORAKS $\mathbf{C}$-matrix is constructed, if the entry in the $m$th column and $n$th row of the matrix $\mathbf{B}$ is nonzero, then the corresponding entry of the matrix $\mathbf{D}$ is required to be zero and vice versa. Note also that the matrix from Eq. (10) can be written as $\mathbf{B} + \mathbf{D}$, while the matrix from Eq. (11) can be written as $\mathbf{B} - \mathbf{D}$.

To avoid additional tedious notation, we will assume in our proof sketch that the rows and columns of the matrix $\mathbf{B}$ have been permuted in such a way that samples from even and odd lines in k-space are never adjacent to one another in the matrix, which is always possible based on the convolutional structure of the LORAKS $\mathbf{C}$-matrix. This allows the $\mathbf{B}$ matrix to be written in a "checkerboard" form

$$\mathbf{B} = \left[\begin{array}{cccccc|cccccc} b_{11}^+ & 0 & b_{13}^+ & 0 & b_{15}^+ & \cdots & 0 & b_{12}^- & 0 & b_{14}^- & 0 & \cdots \\ 0 & b_{22}^+ & 0 & b_{24}^+ & 0 & \cdots & b_{21}^- & 0 & b_{23}^- & 0 & b_{25}^- & \cdots \\ b_{31}^+ & 0 & b_{33}^+ & 0 & b_{35}^+ & \cdots & 0 & b_{32}^- & 0 & b_{34}^- & 0 & \cdots \\ \vdots & \vdots & \vdots & \vdots & \vdots & \ddots & \vdots & \vdots & \vdots & \vdots & \vdots & \ddots \end{array}\right], \quad (\text{S.3})$$

where $b_{ij}^+$ and $b_{ij}^-$ are the nonzero entries of the $\mathbf{B}$ matrix corresponding to positive and negative readout polarities, respectively. Using the same permutation scheme, the matrix $\mathbf{D}$ can similarly be written in the corresponding complementary "checkerboard" form:

$$\mathbf{D} = \left[\begin{array}{cccccc|cccccc} 0 & d_{12}^+ & 0 & d_{14}^+ & 0 & \cdots & d_{11}^- & 0 & d_{13}^- & 0 & d_{15}^- & \cdots \\ d_{21}^+ & 0 & d_{23}^+ & 0 & d_{25}^+ & \cdots & 0 & d_{22}^- & 0 & d_{24}^- & 0 & \cdots \\ 0 & d_{32}^+ & 0 & d_{34}^+ & 0 & \cdots & d_{31}^- & 0 & d_{33}^- & 0 & d_{35}^- & \cdots \\ \vdots & \vdots & \vdots & \vdots & \vdots & \ddots & \vdots & \vdots & \vdots & \vdots & \vdots & \ddots \end{array}\right], \quad (\text{S.4})$$

Consider the diagonal matrix $\mathbf{Q}_1$ which has the same number of columns as $\mathbf{B}$, and whose diagonal entries alternate in sign in a way that follows the non-zero pattern of the first row of $\mathbf{B}$:

$$\mathbf{Q}_1 = \left[\begin{array}{ccccc|ccccc} 1 & 0 & 0 & 0 & 0 & \cdots & 0 & 0 & 0 & 0 & 0 & \cdots \\ 0 & -1 & 0 & 0 & 0 & \cdots & 0 & 0 & 0 & 0 & 0 & \cdots \\ 0 & 0 & 1 & 0 & 0 & \cdots & 0 & 0 & 0 & 0 & 0 & \cdots \\ 0 & 0 & 0 & -1 & 0 & \cdots & 0 & 0 & 0 & 0 & 0 & \cdots \\ 0 & 0 & 0 & 0 & 1 & \cdots & 0 & 0 & 0 & 0 & 0 & \cdots \\ \vdots & \vdots & \vdots & \vdots & \vdots & \ddots & \vdots & \vdots & \vdots & \vdots & \vdots & \ddots \\ 0 & 0 & 0 & 0 & 0 & \cdots & -1 & 0 & 0 & 0 & 0 & \cdots \\ 0 & 0 & 0 & 0 & 0 & \cdots & 0 & 1 & 0 & 0 & 0 & \cdots \\ 0 & 0 & 0 & 0 & 0 & \cdots & 0 & 0 & -1 & 0 & 0 & \cdots \\ 0 & 0 & 0 & 0 & 0 & \cdots & 0 & 0 & 0 & 1 & 0 & \cdots \\ 0 & 0 & 0 & 0 & 0 & \cdots & 0 & 0 & 0 & 0 & -1 & \cdots \\ \vdots & \vdots & \vdots & \vdots & \vdots & \ddots & \vdots & \vdots & \vdots & \vdots & \vdots & \ddots \end{array}\right]. \quad (\text{S.5})$$

Similarly, consider the diagonal matrix $\mathbf{Q}_2$ which has the same number of rows as $\mathbf{B}$, and whose diagonal entries alternate in sign in a way that follows the non-zero pattern of the first column of $\mathbf{B}$:

$$\mathbf{Q}_2 = \begin{bmatrix} 1 & 0 & 0 & \cdots \\ 0 & -1 & 0 & \cdots \\ 0 & 0 & 1 & \cdots \\ \vdots & \vdots & \vdots & \ddots \end{bmatrix}. \quad (\text{S.6})$$

We make the following observations:

- The matrices $\mathbf{Q}_1$ and $\mathbf{Q}_2$ are unitary and satisfy $\mathbf{Q}_1^{-1} = \mathbf{Q}_1$ and $\mathbf{Q}_2^{-1} = \mathbf{Q}_2$.
- The matrix $\mathbf{B}$ is structured in such a way that $\mathbf{B}\mathbf{Q}_1 = \mathbf{Q}_2 \mathbf{B}$.
- The matrix $\mathbf{D}$ is structured in such a way that $\mathbf{D}\mathbf{Q}_1 = -\mathbf{Q}_2 \mathbf{D}$.
- The matrix $\mathbf{Q}_2(\mathbf{B} + \mathbf{D})\mathbf{Q}_1$ simplifies according to

$$\begin{aligned} \mathbf{Q}_2(\mathbf{B}+\mathbf{D})\mathbf{Q}_1 &= \mathbf{Q}_2\mathbf{Q}_2(\mathbf{B}-\mathbf{D}) \\ &= \mathbf{B} - \mathbf{D}. \end{aligned} \quad (\text{S.7})$$

From Eq. (S.7), we can infer that if we write the singular value decomposition of $\mathbf{B} + \mathbf{D}$ as $\mathbf{B}+\mathbf{D} = \mathbf{U}\boldsymbol{\Sigma}\mathbf{V}^H$, then the matrix $\mathbf{B} - \mathbf{D}$ can be written as $\tilde{\mathbf{U}}\boldsymbol{\Sigma}\tilde{\mathbf{V}}^H$, where $\tilde{\mathbf{U}} = \mathbf{Q}_2 \mathbf{U}$ and $\tilde{\mathbf{V}} = \mathbf{Q}_1 \mathbf{V}$. Since $\tilde{\mathbf{U}}$ and $\tilde{\mathbf{V}}$ are matrices with orthonormal columns, we must have that $\tilde{\mathbf{U}}\boldsymbol{\Sigma}\tilde{\mathbf{V}}^H$ is a valid singular value decomposition of $\mathbf{B} - \mathbf{D}$. Thus, we can conclude that $\mathbf{B} - \mathbf{D}$ and $\mathbf{B} + \mathbf{D}$ have identical singular values. This completes the proof of the theorem. □

*Proof of Corollary 1*

Assume that the vectors **y** and **z** are chosen such that the expressions in Eq. (9) represent an optimal solution to Eq. (5). Theorem 1 then tells us that the vectors from Eq. (12) represent another optimal solution to Eq. (5). These two solutions are identical to one another if and only if $\mathbf{A}_-^H \mathbf{A}_- \mathbf{y} = \mathbf{0}$ and $\mathbf{A}_+^H \mathbf{A}_+ \mathbf{z} = \mathbf{0}$, in which case both solutions are equal to the zero-filled solution. In this case, the zero-filled solution is clearly an optimal solution to Eq. (5), and must be the unique optimal solution if Eq. (5) only has a single solution. If either $\mathbf{A}_-^H \mathbf{A}_- \mathbf{y} \neq \mathbf{0}$ or $\mathbf{A}_+^H \mathbf{A}_+ \mathbf{z} \neq \mathbf{0}$, then Eqs. (9) and (12) represent two distinct solutions to Eq. (5), indicating that Eq. (5) does not have a unique solution. □

*Proof of Corollary 2*

Based on the proof of Corollary 1, we know that if Eq. (5) has a solution that is not equal to the zero-filled measured data, then we can use Eqs. (9) and (12) to obtain a pair of two distinct solutions to Eq. (5). Let $\mathbf{k}_1^\pm$ and $\mathbf{k}_2^\pm$ denote these two solutions, and notice that by the definition of an optimal solution, we must have that $L_\mathbf{C}(\mathbf{k}_1^\pm) = L_\mathbf{C}(\mathbf{k}_2^\pm)$ and that $L_\mathbf{C}(\mathbf{k}_1^\pm) \leq L_\mathbf{C}(\mathbf{k})$ for all possible candidate solutions $\mathbf{k}$.

Corollary 2 is easily proven based on the definition of a convex function. Specifically, if $L_\mathbf{C}(\mathbf{y})$ is convex, then it must satisfy [1]

$$L_\mathbf{C}(\alpha \mathbf{y}_1 + (1-\alpha)\mathbf{y}_2) \leq \alpha L_\mathbf{C}(\mathbf{y}_1) + (1-\alpha) L_\mathbf{C}(\mathbf{y}_2), \quad \text{(S.8)}$$

for every possible pair of vectors $\mathbf{y}_1$ and $\mathbf{y}_2$ and for every real-valued scalar $\alpha$ between 0 and 1.

Setting $\mathbf{y}_1 = \mathbf{k}_1^\pm$ and $\mathbf{y}_2 = \mathbf{k}_2^\pm$ in Eq. (S.8) leads to

$$\begin{aligned} L_\mathbf{C}\left(\alpha \mathbf{k}_1^\pm + (1-\alpha)\mathbf{k}_2^\pm\right) &\leq \alpha L_\mathbf{C}\left(\mathbf{k}_1^\pm\right) + (1-\alpha) L_\mathbf{C}\left(\mathbf{k}_2^\pm\right) \\ &= L_\mathbf{C}\left(\mathbf{k}_1^\pm\right), \end{aligned} \quad \text{(S.9)}$$

Combining Eq. (S.9) with the previous observation that $L_\mathbf{C}(\mathbf{k}_1^\pm) \leq L_\mathbf{C}(\mathbf{k})$ for all possible candidate solutions $\mathbf{k}$ implies that $L_\mathbf{C}\left(\alpha \mathbf{k}_1^\pm + (1-\alpha)\mathbf{k}_2^\pm\right) = L_\mathbf{C}\left(\mathbf{k}_1^\pm\right)$. As a result, $\alpha \mathbf{k}_1^\pm + (1-\alpha)\mathbf{k}_2^\pm$ must also be an optimal solution of Eq. (5) for every possible choice of $0 \leq \alpha \leq 1$, and we have successfully proven that there exist an infinite number of solutions.

Specifically, an infinite set of solutions is given by $\{\mathbf{k}_\alpha^\pm \equiv \alpha \mathbf{k}_1^\pm + (1-\alpha)\mathbf{k}_2^\pm \mid \alpha \in [0,1]\}$. Additionally, the zero-filled solution is obtained as one of these solutions, corresponding to the specific choice of $\alpha = 0.5$. □

## S.II. Experimental Details

### A. In vivo data

The *in vivo* data shown in Figs. 1, 4, 5, 7, and 9(b) of the main paper and supplementary Figs. S1-S9 was acquired using a gradient echo EPI sequence on a Siemens (Erlangen, Germany) 3T Prisma Fit using a standard product 32-channel receiver array. The imaging parameters were: FOV = 220 mm × 220 mm; matrix size 128×128; slice thickness = 3 mm; TR = 2.08 sec; TE = 47 msec; acceleration factor $R \in \{1, 2, 3, 4, 5\}$.

### B. Phantom data

The phantom data shown in Fig. 6 was acquired using a gradient echo EPI sequence on a Siemens 3T Tim Trio using a standard product 12-channel receiver array (compressed in hardware down to 4 channels). The imaging parameters were: FOV = 24 mm × 24 mm; matrix size 64×64; slice thickness = 5 mm; TR = 1.46 sec; TE = 38 msec.

The phantom data used in Figs. 8 and 9(a) was acquired using a gradient echo EPI sequence on a Siemens 3T Tim Trio using a standard product 12-channel receiver array. The imaging parameters were: FOV = 24 mm × 24 mm; matrix size 96×96; slice thickness = 5 mm; TR = 60 msec; TE = 28 msec.

## S.III. Majorize-Minimize Algorithm

For the sake of completeness, this section provides an operational description of the majorize-minimize algorithm we use to solve Eq. (13). The key observation [2], [3] that enables this algorithm is that the function $J_r(\mathbf{G})$ from Eq. (8) is majorized at a point $\hat{\mathbf{G}}^{(j)}$ by the function

$$g\left(\mathbf{G}, \hat{\mathbf{G}}\right) = \|\mathbf{G} - \hat{\mathbf{G}}^{(j)}\|_F^2. \quad \text{(S.10)}$$

Using this majorant, Eq. (13) can be optimized by iteratively solving easy-to-optimize surrogate problems. Specifically, using $\hat{\boldsymbol{\rho}}^{(j)}$ to denote the estimate of $\{\boldsymbol{\rho}^+, \boldsymbol{\rho}^-\}$ at the $j$th iteration, and given some initial guess $\hat{\boldsymbol{\rho}}^{(0)}$, we minimize Eq. (13) by iteratively solving the surrogate problems

$$\begin{aligned} \hat{\boldsymbol{\rho}}^{(j+1)} = \arg\min_{\{\boldsymbol{\rho}^+, \boldsymbol{\rho}^-\}} &\|\mathbf{E}_+ \boldsymbol{\rho}^+ - \mathbf{d}_{\text{tot}}^+\|_2^2 \\ &+ \|\mathbf{E}_- \boldsymbol{\rho}^- - \mathbf{d}_{\text{tot}}^-\|_2^2 \\ &+ \lambda \|\begin{bmatrix} \mathbf{S}_P(\mathbf{E}_+ \boldsymbol{\rho}^+) & \mathbf{S}_P(\mathbf{E}_- \boldsymbol{\rho}^-) \end{bmatrix} - \mathbf{G}_r^{(j)}\|_F^2 \end{aligned} \quad \text{(S.11)}$$

for $j=0, 1, 2, \ldots$ until convergence, where $\mathbf{G}_r^{(j)}$ is the optimal rank-$r$ approximation (obtained by truncating the singular value decomposition) of the matrix formed by numerically evaluating $\begin{bmatrix} \mathbf{S}_P(\mathbf{E}_+ \boldsymbol{\rho}^+) & \mathbf{S}_P(\mathbf{E}_- \boldsymbol{\rho}^-) \end{bmatrix}$ at the point $\{\boldsymbol{\rho}^+, \boldsymbol{\rho}^-\} = \hat{\boldsymbol{\rho}}^{(j)}$. These surrogate problems have the form of simple linear least squares problems, and can be solved using standard iterative algorithms like the conjugate gradient algorithm. See Refs. [2], [3] for a more detailed description and justification of this algorithmic approach.

## S.IV. Supplementary Figures and Tables

Supplementary figures that accompany the main paper are shown over the following several pages. See the main paper for additional description.

## References

[1] D. G. Luenberger, *Optimization by Vector Space Methods.* New York: John Wiley & Sons, 1969.
[2] J. P. Haldar, "Low-Rank Modeling of Local-Space Neighborhoods (LORAKS) for Constrained MRI," *IEEE Trans. Med. Imag.*, vol. 33, pp. 668–681, 2014.
[3] J. P. Haldar, "Low-rank modeling of local k-space neighborhoods (LORAKS): Implementation and examples for reproducible research," University of Southern California, Los Angeles, CA, Tech. Rep. USC-SIPI-414, April 2014, http://sipi.usc.edu/reports/abstracts.php?rid=sipi-414.

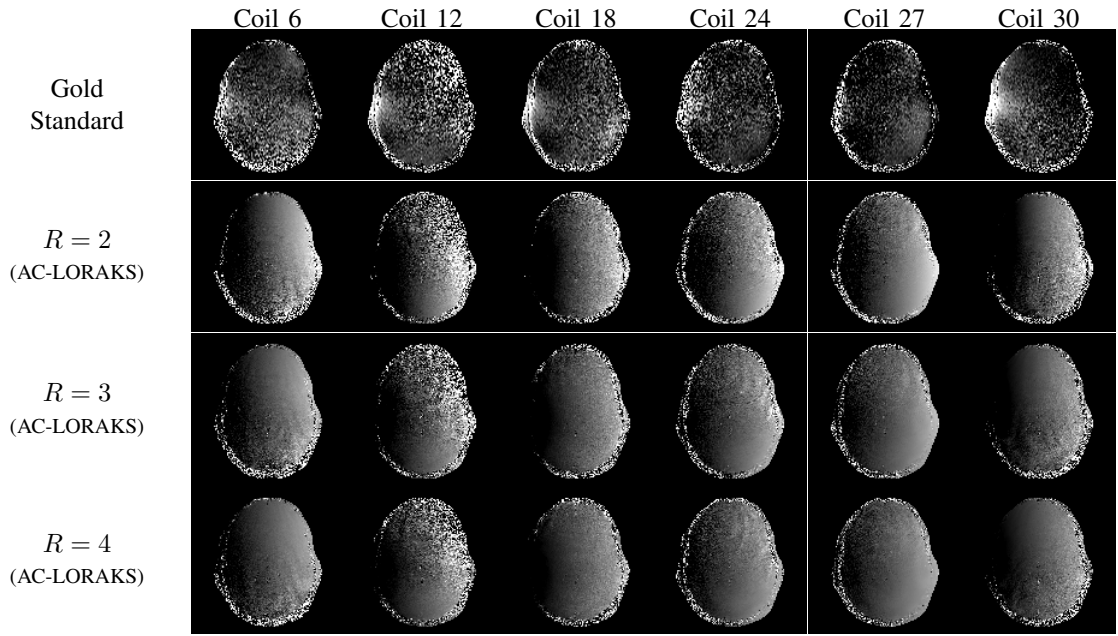

Fig. S1: Phase differences between RO$^+$ and RO$^-$ for a representative subset of coils from the prospectively undersampled data shown in Fig. 4. In cases with accelerated data (where no gold standard is available), we show the phase difference obtained after AC-LORAKS reconstruction. Unlike most other phase images shown in this paper (which do not include background masking and show the entire phase range from $-\pi$ to $\pi$), we have taken steps to make the phase images in this figure easier to visualize. In particular, we have masked the background noise to make it easier to focus on the signal structure. In addition, we have shown a restricted phase range for both the gold standard (black = $0.95\pi$ radians, white = $1.02\pi$ radians) and the reconstructions using AC-LORAKS (black = $0.89\pi$ radians, white = $1.21\pi$ radians).

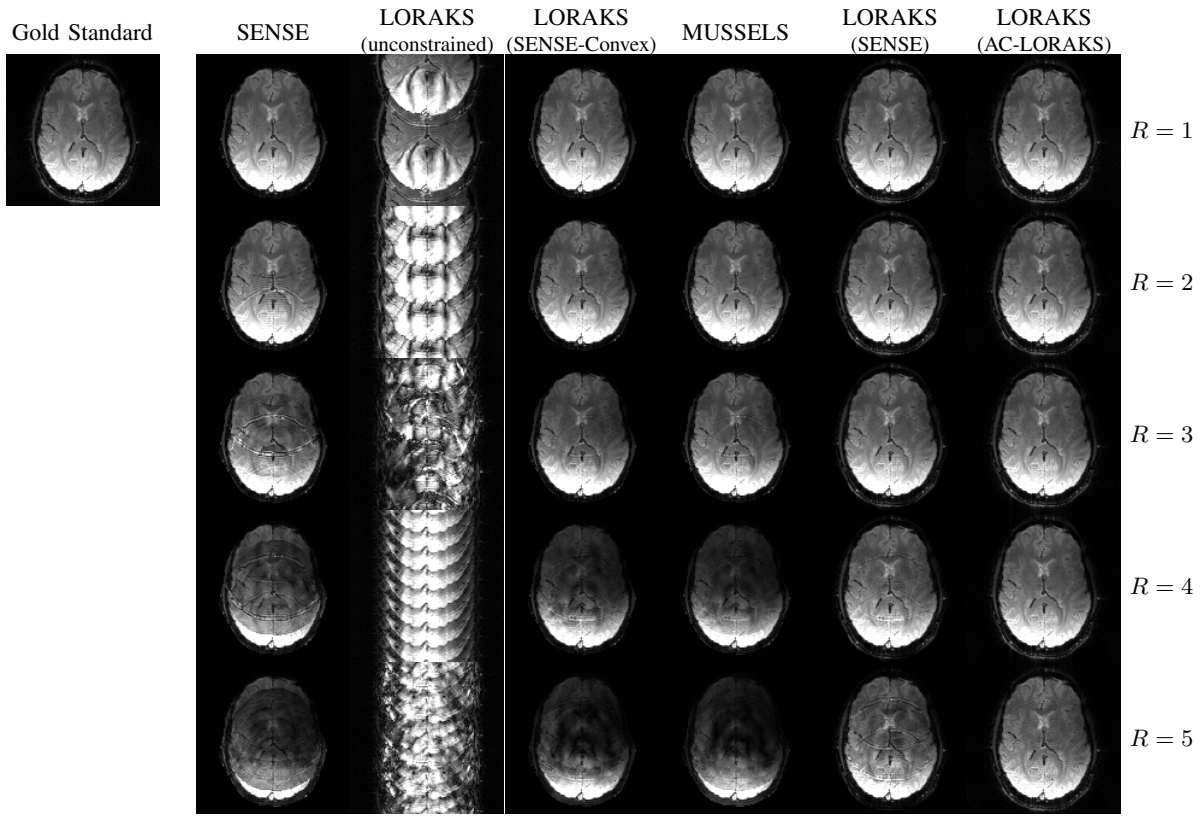

Fig. S2: Comparison of different reconstruction techniques using retrospectively undersampled *in vivo* single-shot EPI data to simulate different parallel imaging acceleration factors.

TABLE S1: NRMSE values for the images shown in Fig. S2.

| $R$ | SENSE | LORAKS (unconstr) | LORAKS (SENSE-Conv) | MUSSELS | LORAKS (SENSE) | LORAKS (AC) |
|---|---|---|---|---|---|---|
| 1 | 0.17 | 0.70 | 0.17 | 0.17 | 0.09 | **0.03** |
| 2 | 0.23 | 0.86 | 0.20 | 0.20 | 0.10 | **0.05** |
| 3 | 0.38 | 1.04 | 0.27 | 0.26 | 0.11 | **0.06** |
| 4 | 0.53 | 0.94 | 0.42 | 0.49 | 0.15 | **0.07** |
| 5 | 0.65 | 1.00 | 0.62 | 0.67 | 0.25 | **0.12** |

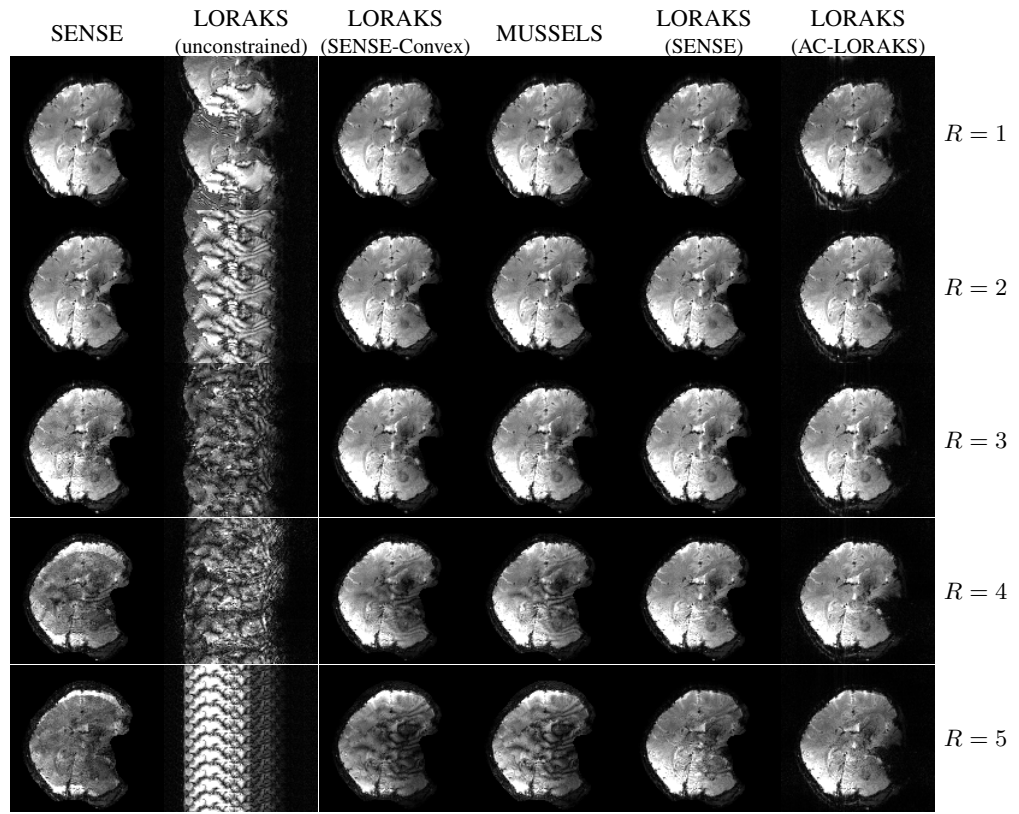

Fig. S3: Comparison of different reconstruction techniques using prospectively undersampled *in vivo* double-oblique single-shot EPI data at different parallel imaging acceleration factors.

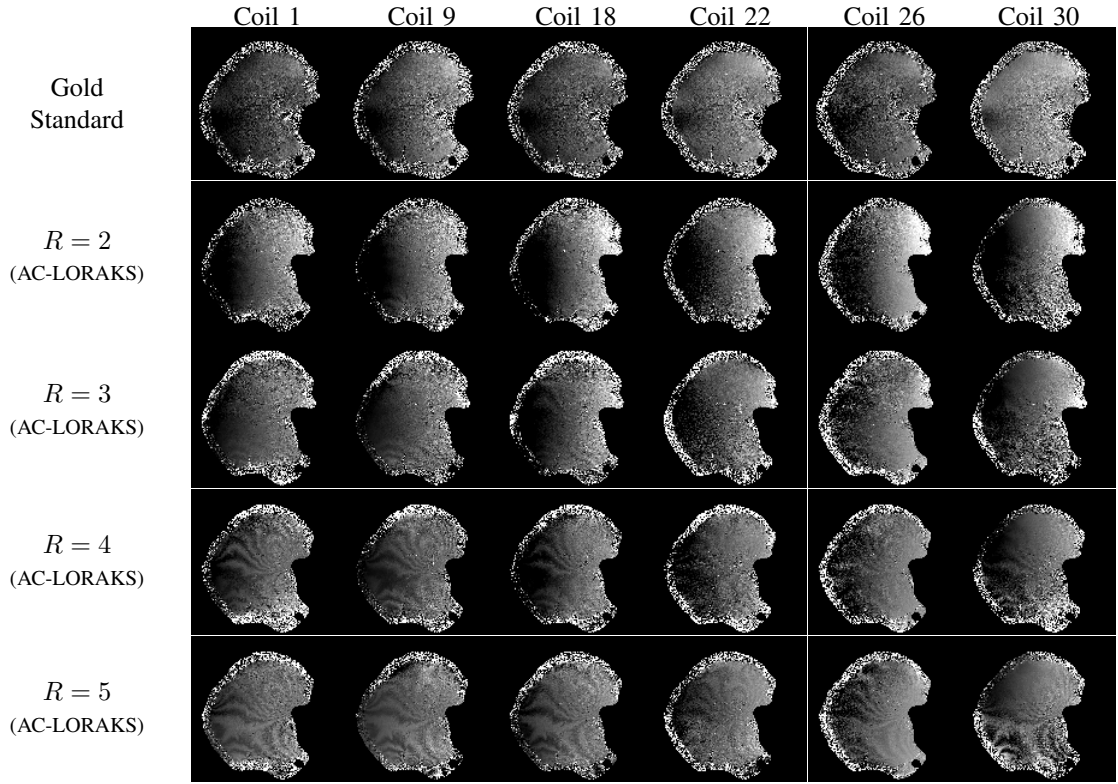

Fig. S4: Phase differences between RO$^+$ and RO$^-$ for a representative subset of coils from the prospectively undersampled double-oblique data shown in Fig. S3. In cases with accelerated data (where no gold standard is available), we show the phase difference obtained after reconstruction using LORAKS with AC-LORAKS constraints. Unlike most other phase images shown in this paper (which do not include background masking and show the entire phase range from $-\pi$ to $\pi$), we have taken steps to make the phase images in this figure easier to visualize. In particular, we have masked the background noise to make it easier to focus on the signal structure. In addition, we have shown a restricted phase range for both the gold standard (black = $0.95\pi$ radians, white = $1.02\pi$ radians) and the reconstructions using AC-LORAKS (black = $0.86\pi$ radians, white = $1.18\pi$ radians).

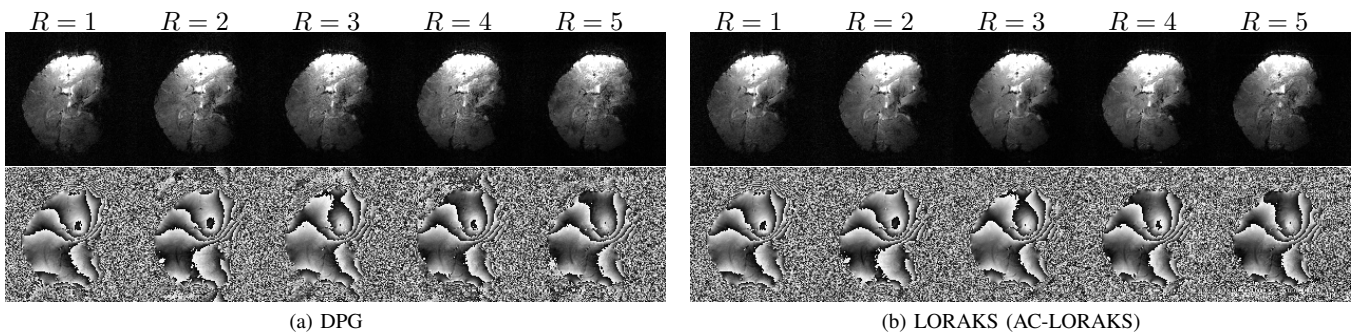

Fig. S5: Comparison between (a) DPG and (b) LORAKS with AC-LORAKS constraints for the real single-shot EPI *in vivo* brain double-oblique data from Fig. S3. Instead of showing coil-combined images, a single representative channel is shown to avoid contamination of the phase characteristics induced by coil combination.

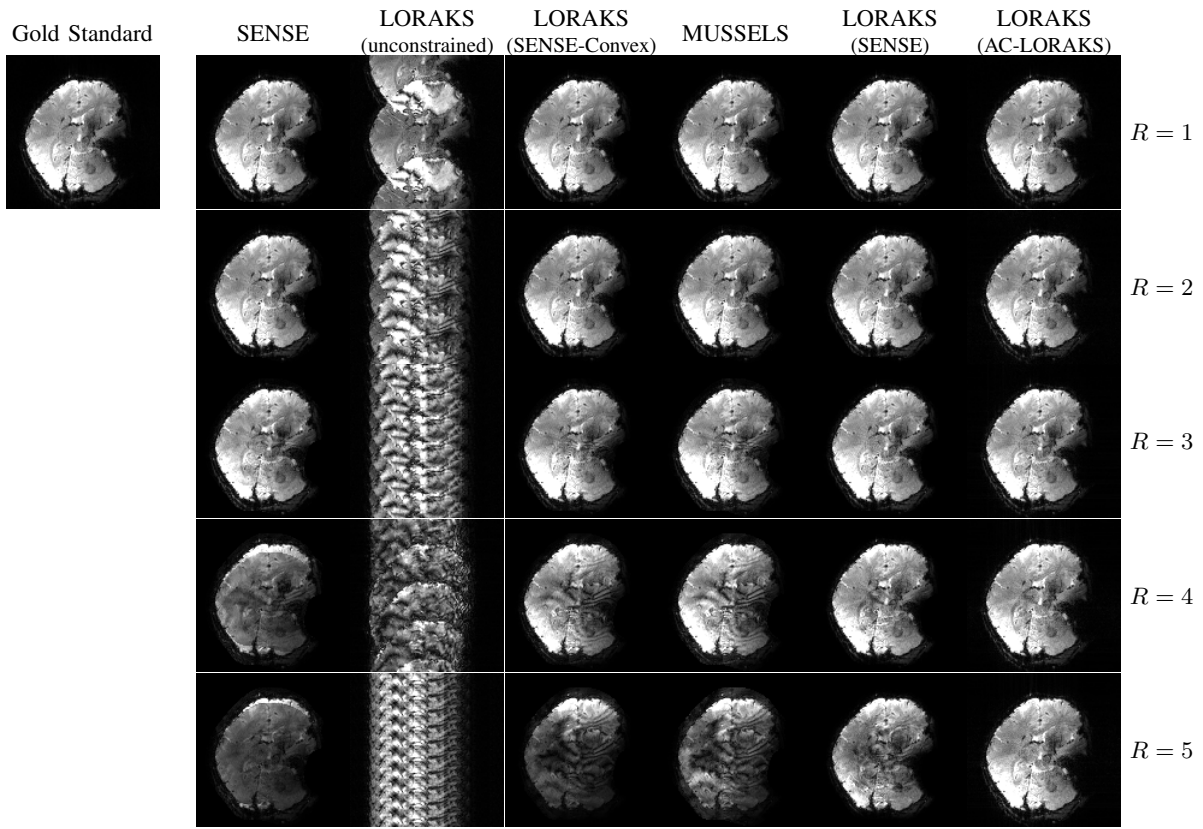

Fig. S6: Comparison of different reconstruction techniques using retrospectively undersampled *in vivo* double-oblique data to simulate single-shot EPI experiments at different parallel imaging acceleration factors.

TABLE S2: NRMSE values for the images shown in Fig. S6.

| $R$ | SENSE | LORAKS (unconstr) | LORAKS (SENSE-Conv) | MUSSELS | LORAKS (SENSE) | LORAKS (AC) |
|---|---|---|---|---|---|---|
| 1 | 0.06 | 0.70 | 0.09 | 0.09 | 0.07 | **0.03** |
| 2 | 0.07 | 0.86 | 0.13 | 0.15 | 0.08 | **0.07** |
| 3 | 0.19 | 0.91 | 0.20 | 0.25 | 0.11 | **0.10** |
| 4 | 0.46 | 0.99 | 0.35 | 0.40 | 0.18 | **0.12** |
| 5 | 0.61 | 0.95 | 0.65 | 0.74 | 0.29 | **0.11** |

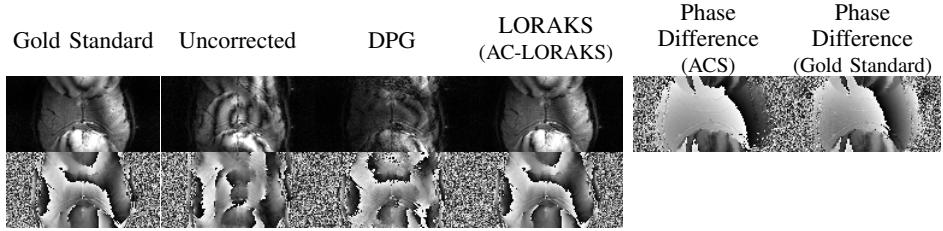

Fig. S7: (top-left) Magnitude and (bottom-left) phase images corresponding to reconstruction of unaccelerated ($R = 1$) *in vivo* axial multi-channel single-shot EPI data acquired with a small FOV, and with extra simulated nonlinear phase differences between $RO^+$ and $RO^-$. The images on the right show the phase differences that were used between $RO^+$ and $RO^-$, which were set differently for the ACS data (used for calibration) than they were for the gold standard (used for reconstruction).

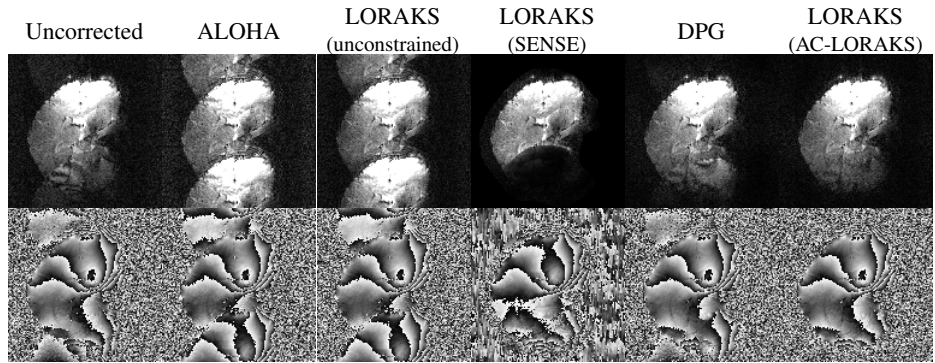

Fig. S8: (top) Magnitude and (bottom) phase images corresponding to reconstruction of unaccelerated ($R = 1$) *in vivo* double-oblique single-channel single-shot EPI data.

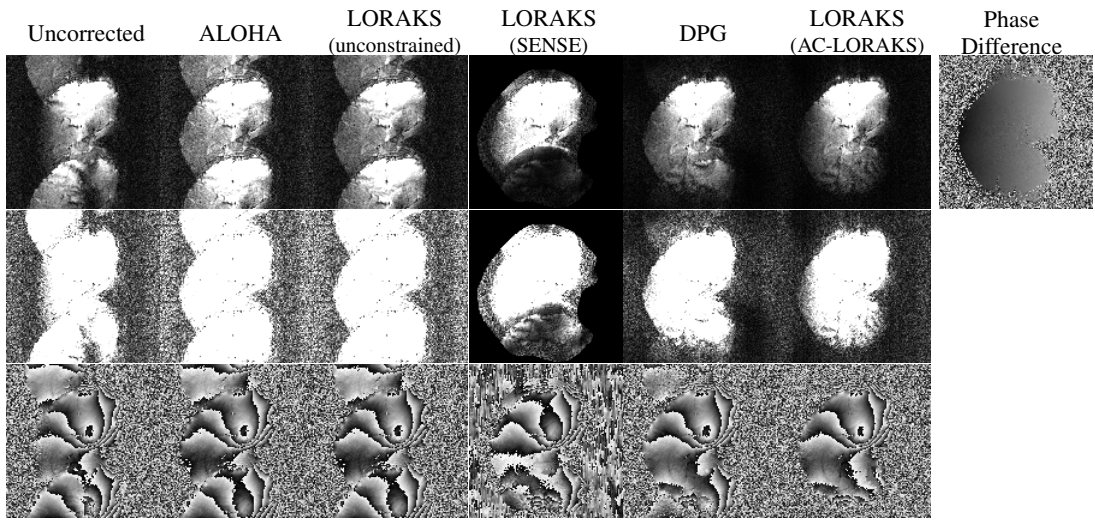

Fig. S9: (top, middle) Magnitude and (bottom) phase images corresponding to reconstruction of unaccelerated ($R = 1$) *in vivo* double-oblique single-channel single-shot EPI with strong simulated phase differences between $RO^+$ and $RO^-$. The middle row shows a different windowing of the magnitude images to more clearly visualize ghost-artifacts. The top-right image shows the simulated phase difference between $RO^+$ and $RO^-$.